\documentclass{article}

\usepackage{iclr2027_conference,times}
\usepackage[T1]{fontenc}
\usepackage[utf8]{inputenc}
\usepackage{amsmath,amssymb,mathtools}
\usepackage{graphicx}
\usepackage{float}
\usepackage{wrapfig}
\usepackage{needspace}
\usepackage{enumitem}
\usepackage{longtable}
\usepackage{booktabs}
\usepackage{array}
\usepackage{tabularx}
\usepackage{etoolbox}
\usepackage[final]{microtype}
\usepackage{hyperref}
\usepackage{url}
\usepackage{xcolor}
\usepackage{colortbl}
\usepackage{multirow}
\definecolor{citeblue}{RGB}{31,78,168}
\hypersetup{
  colorlinks=true,
  citecolor=citeblue,
  linkcolor=citeblue,
  urlcolor=citeblue,
  pdftitle={Revision-Aware Independent Agent Graphs for Dynamic Reasoning},
  pdfauthor={Yan Luo, Selim-Antoine Lali, Jeremy Moebel, Iliass Khoutaibi, Ahmadou Aidara, Mengyu Wang},
  pdfsubject={Artificial intelligence; multi-agent reasoning; dynamic task routing}
}

\newcolumntype{R}{>{\raggedleft\arraybackslash}p{0.50in}}
\AtBeginEnvironment{longtable}{\scriptsize\setlength{\tabcolsep}{2pt}}

\title{Revision-Aware Independent Agent Graphs for Dynamic Reasoning}
\date{}

\author{
Yan Luo\thanks{Contributed equally as co-first authors.}\\
Harvard AI and Robotics Lab\\
Harvard University\\
Boston, MA, USA
\And
Selim-Antoine Lali\footnotemark[1]\\
Harvard AI and Robotics Lab\\
Harvard University\\
Boston, MA, USA
\And
Jeremy Moebel\footnotemark[1]\\
Harvard AI and Robotics Lab\\
Harvard University\\
Boston, MA, USA
\AND
Iliass Khoutaibi\\
Harvard AI and Robotics Lab\\
Harvard University\\
Boston, MA, USA
\And
Ahmadou Aidara\\
Harvard AI and Robotics Lab\\
Harvard University\\
Boston, MA, USA
\And
Mengyu Wang\\
Harvard AI and Robotics Lab\\
Harvard University\\
Boston, MA, USA
}

\iclrfinalcopy

\begin{document}
\raggedbottom

\maketitle
\fancyhead[L]{Preprint}

\begin{abstract}
Conventional reasoning protocols present a fixed, preselected task, so they cannot test whether an agent propagates relevant updates, preserves unaffected work, or reconstructs a historical task binding.  We therefore study \emph{dynamic task routing}, in which an event stream revises task bindings and a system must select the document version valid at each query time before solving it.  To study this problem, we repurpose six widely used benchmarks: MMLU, MMLU-Pro, MedMCQA, MATH, GPQA, and HumanEval into 31{,}119 dynamic episodes comprising 373{,}428 temporally categorized queries. This setting exposes a central trade-off: recomputing after every event wastes work, whereas unguarded reuse returns stale conclusions.  We introduce the Revision-Aware Independent Agent Graph (RIAG), a bounded multi-agent policy that separates deterministic temporal resolution from task reasoning.  RIAG caches solutions by immutable document identity, starts each fresh task with two unexposed attempts, and conditionally invokes audit and repair, using at most four calls per document version.  On this collection, homogeneous RIAG achieves 54.24\% joint routing-and-answer accuracy at 0.62 calls/query, compared with 32.22\% at 18.00 calls/query for the strongest comparison method; heterogeneous RIAG reaches 49.78\% at 0.63 calls/query. 

\end{abstract}

\section{Introduction}
\label{sec:introduction}

Language-model agents are being developed as persistent systems that remember
across conversations, accumulate experience in interactive environments, and
act through software interfaces rather than answer a single isolated prompt
\citep{park2023generativeagents,packer2023memgpt,wang2024voyager,yang2024sweagent}.
This shift is practically useful: an agent can carry work across sessions,
incorporate environmental or execution feedback, and update a plan as new
information arrives.  It also makes the agent's input an evolving event stream.
A report can arrive late and describe an earlier state, an instruction can be
visible before it takes effect, and a retraction can change what should have
been believed about the past.  A multi-agent reasoner must therefore decide
which task is currently valid before solving it, whether an earlier solution
still applies after a revision, and whether matching answers are independent
corroboration or merely agreement induced by communication.  Getting the first
decision wrong means reasoning about a superseded task; recomputing after every
event avoids stale reuse but wastes calls and latency on unchanged inputs; and
counting an exposed agent's agreement as a fresh vote overstates the evidence.
These coupled choices motivate our central question: \textit{when an event stream revises which task is active, what should a multi-agent reasoner recompute, what may it reuse, and whose agreement should count as evidence?}

Standard reasoning benchmarks present each model with an already selected,
fixed problem instance
\citep{hendrycks2021mmlu,wang2024mmlupro,pal2022medmcqa,hendrycks2021math,
rein2024gpqa,chen2021humaneval}, and multi-agent evaluations built on such
benchmarks study how agents solve that fixed task through debate, refinement,
aggregation, or graph-structured communication
\citep{du2023multiagentdebate,madaan2023selfrefine,wang2024mixtureofagents,
yun2026graphofagents}.  Broader agent benchmarks add multi-turn interaction and
partially observable environments \citep{liu2024agentbench,ma2024agentboard},
but evaluate progress or final success on a supplied task rather than whether
the identity of that task remains valid as its governing events are revised.
These settings therefore do not isolate whether a system can distinguish an
event's arrival time from its effective time, propagate a genuine routing
change, ignore an irrelevant or not-yet-effective event, revise a historical
binding after a retraction, or reuse reasoning when a previously solved task
becomes active again.  We define \emph{dynamic task routing} to expose these
capabilities: given only the event prefix received so far, a system must select
the document in force at the requested time and solve it.  We instantiate the
problem by repurposing six widely used benchmarks: MMLU, MMLU-Pro, MedMCQA,
MATH, GPQA, and HumanEval, as dynamic task routing datasets while preserving
their original questions and evaluators. 

\begin{figure}[t]
  \centering
  \includegraphics[width=\textwidth]{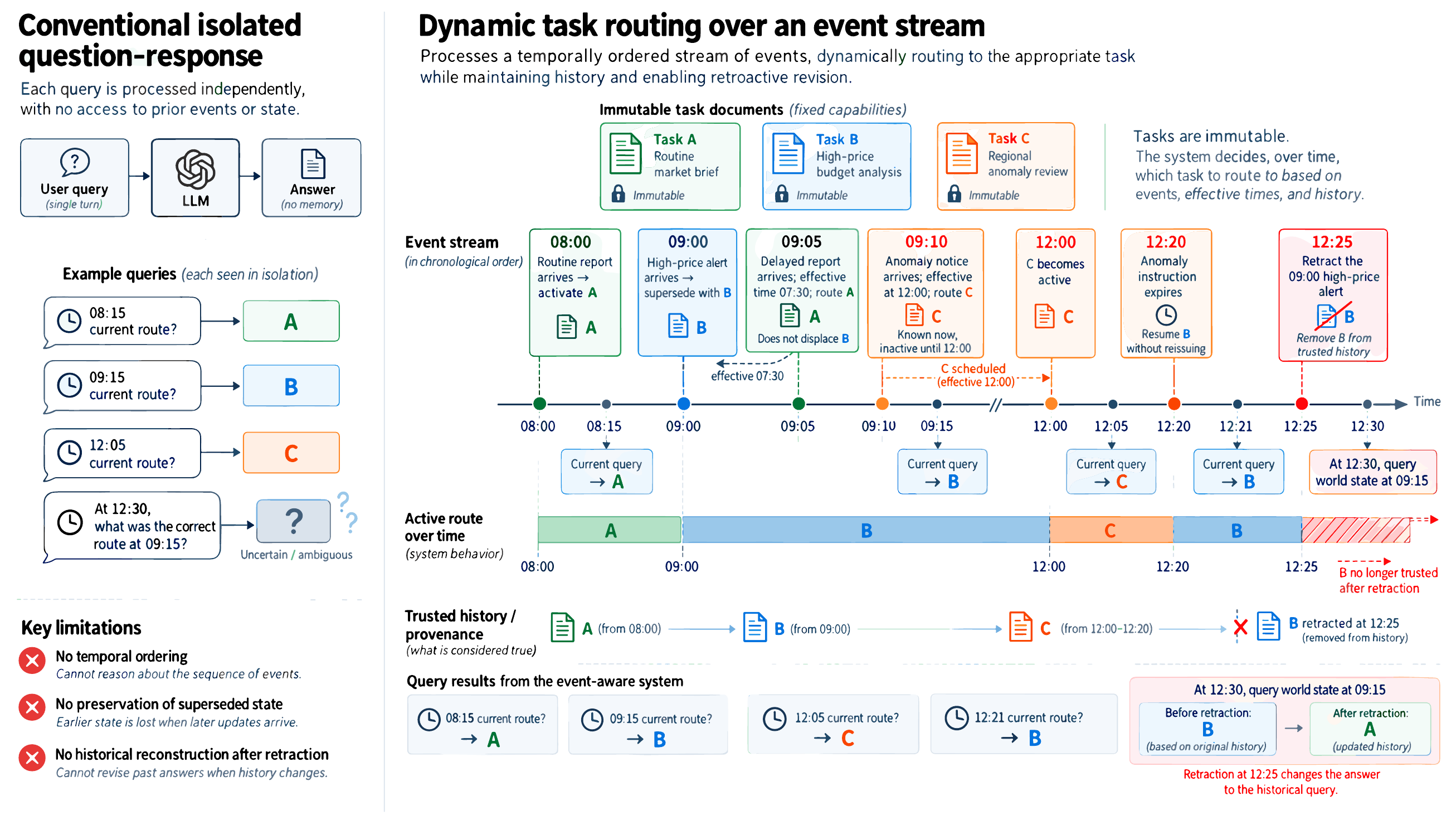}
  \caption{Conventional isolated question and response protocol versus dynamic task
  routing protocol. The dynamic protocol retains an arrival ordered event history,
  distinguishes arrival time from effective time, routes each query to a
  fixed-content task version, restores an underlying route after expiration, and
  revises historical resolution when a previously trusted event is retracted.}
  \label{fig:routing-teaser}
\end{figure}

Figure~\ref{fig:routing-teaser} illustrates the resulting shift in evaluation.
Conventional evaluation supplies one fixed task and scores one response.  In
dynamic task routing, we hold task content fixed as an experimental control
while an arrival-ordered event history changes which document is active.  This
separates temporal binding and reuse from content editing; a content change
would instead create a new immutable document version and force recomputation.
A delayed report or future-effective instruction should leave the current route
unchanged, an expiration can restore an earlier route, and a retraction can
revise the document selected for a historical query.  The system must therefore
resolve the valid version before deciding whether an earlier solution is
reusable.





We introduce the \emph{Revision-Aware Independent
Agent Graph} (RIAG), a bounded inference policy that composes a dynamic name
resolver with a content addressed reasoner (Section~\ref{sec:method}) and
answers the three questions directly.  \textbf{Which task?}  At every query, a
deterministic resolver uses only the visible event prefix to select the
task-document version valid at the requested time.  \textbf{Recompute or
reuse?}  RIAG caches each solution under a key containing the document version
and inference configuration, and reuses the solution only when that key matches
exactly, so a cache hit costs no model call.  \textbf{Whose agreement counts?}
On a cache miss, a solver and a checker answer without seeing each other's
output, so matching proposals provide independent corroboration.  Only when
they disagree or a check fails does RIAG call a targeted auditor and, if
needed, one repairer, which limits a fresh task to four calls.  Every invocation
records which earlier outputs it received, distinguishing independent agreement
from agreement after communication.  RIAG requires no training or learned
router and uses only a black box model interface.

We compare RIAG with Debate \citep{du2023multiagentdebate}, Self-Consistency
\citep{wang2023selfconsistency}, Refine \citep{madaan2023selfrefine}, ReConcile
\citep{chen2024reconcile}, Mixture-of-Agents \citep{wang2024mixtureofagents},
Self-MoA \citep{li2025selfmoa}, an answer generation adaptation of CIA
\citep{wu2026cia}, and Graph-of-Agents with max and mean pooling
\citep{yun2026graphofagents}.  Across the full collection of 31,119 episodes and
373,428 scheduled queries per method, RIAG reaches 54.24\% joint
accuracy, compared with 32.22\% for the strongest comparison method, Debate,
while using 0.62 rather than 18.00 model calls per query.  It selects the
correct document on every query and outperforms the strongest conventional
baseline on each of the six source datasets.  Its joint accuracy also remains
between 53.89\% and 54.50\% across baseline, changed, propagation, preservation,
and historical queries, whereas every conventional method loses between 24 and
29 percentage points from baseline to historical queries.  On MATH, 76.9\% of
queries are answered from the cache and require no new model call.  Assigning
different specialized models across RIAG's roles makes the initial solver and
checker responses more diverse, yet final accuracy falls from 64.13\% with one
shared model to 54.88\% with the specialized model pool.  Together, these
results indicate that reliable temporal
binding and selective reuse can matter more than either repeated deliberation
or model diversity alone.



Our contributions are as follows:
\begin{itemize}[leftmargin=*, topsep=0pt, partopsep=0pt, before=\vspace{-3pt}]
  \item We introduce dynamic task routing as an evaluation setting for agents
  operating over revisable event histories.  We transform six widely used
  benchmarks into 31,119 episodes and 373,428 checkpoint queries that test
  propagation, preservation, historical revision, and reuse, while scoring
  document selection, answer correctness, and their conjunction separately.
  \item We propose RIAG, which combines deterministic temporal resolution,
  exact reuse by document version and inference configuration, and an exposure
  aware invocation graph.  Two agents commit independently before any exchange,
  and audit and repair are invoked only when needed, with at most four calls for
  a fresh task and no call for a cache hit.
  \item We provide a broad empirical and mechanistic study across homogeneous
  and specialized model pools and eight comparison families.  The analysis
  connects RIAG's gains to accurate task binding and temporal reuse, evaluates
  sensitivity to inference settings, and shows that greater response diversity
  does not necessarily improve final accuracy.
\end{itemize}

\section{Related Work}
\label{sec:related-work}

\paragraph{Graph-structured agent orchestration.}

Several systems represent language-model workflows as graphs whose nodes are
agents or operations and whose edges carry intermediate outputs. GPTSwarm
optimizes node prompts and communication edges \citep{zhuge2024gptswarm}, while
AFlow searches over code-represented workflows using execution feedback
\citep{zhang2025aflow}. Graph-of-Agents develops a graph-based collaboration
framework and evaluates it on isolated questions from the six benchmarks that
our dynamic task routing episodes are built from \citep{yun2026graphofagents}. Our work uses a graph for a different purpose: its
vertices are concrete model invocations and its edges record which outputs were
exposed to later calls. The graph is therefore an auditable execution trace that
shows which evidence each answer depended on. Our work further combines this
trace with temporal document resolution, version-keyed caching, and a fixed
four-call budget.



\paragraph{Multi-agent deliberation and specialization.}

Multi-agent LLM systems improve test-time reasoning by eliciting multiple
candidate solutions and exchanging them through structured interaction.
Multi-agent debate iteratively shares arguments among model instances
\citep{du2023multiagentdebate}; ReConcile combines heterogeneous agents through
round-table discussion and confidence-weighted consensus
\citep{chen2024reconcile}; and Mixture-of-Agents aggregates outputs through
successive agent layers \citep{wang2024mixtureofagents}.
These systems
primarily organize
communication for a task that has already been selected. Our work additionally
resolves which task is valid under an evolving event history. It preserves two
unexposed initial proposals, treats post-exposure agreement as dependent
evidence, and opens audit and repair stages only after disagreement or a
structural failure.

\paragraph{Agent memory, reflection, and revision.}

Long-lived agents require mechanisms for retaining and revising prior state.
Generative Agents stores experience, synthesizes reflections, and retrieves
memories for later planning \citep{park2023generativeagents}. Reflexion places
verbal feedback in episodic memory to improve subsequent attempts
\citep{shinn2023reflexion}, while MemGPT manages multiple memory tiers to extend
effective context across long interactions \citep{packer2023memgpt}. RIAG reuses a result
only when its cache key matches exactly: the same task document, model
assignment, prompt policy, seed, and decoding configuration within the
episode. It first
decides which document is in force and only then looks up the cache, so a
route that returns to an earlier document reuses its answer, while an answer is
never carried over to a different document.

\flushbottom
\section{Dynamic Task Routing Protocol}
\label{sec:routing-data}

Conventional isolated question and response evaluation presents one task, requests
one answer, and resets the interaction.  It measures whether a model can solve
the supplied task, but not whether the model can identify which task is valid
after a sequence of delayed, scheduled, superseded, expired, or retracted
instructions.  We therefore turn standard benchmark questions into
\emph{dynamic task routing} episodes, in which each response names the
selected document and its answer.

\paragraph{Illustrative example.}
As shown in Figure~\ref{fig:routing-teaser}, a fleet operator has three
fixed-content task documents: task A requests
a routine market brief, task B requests a high-price budget analysis, and task
C requests a regional anomaly review.  The event stream and its correct
interpretation are as follows.
\begin{enumerate}
  \setlength{\itemsep}{1pt}
  \setlength{\parskip}{0pt}
  \item At 08:00, a routine report activates task A.  A current query at 08:15
  must therefore return A and solve the market-brief document.
  \item At 09:00, a high-price alert supersedes the routine route with task B.
  A current query at 09:15 must return B and solve the budget-analysis document.
  \item At 09:05, the system receives a delayed report whose effective time is
  07:30 and whose route is A.  Although this report arrives after the high-price
  alert, its earlier effective time means that it must not displace B.
  \item At 09:10, an anomaly notice assigning task C arrives with an effective
  time of 12:00.  It is known to the system immediately but must not change the
  active route before noon.  Once noon passes, a current query at 12:05 routes
  to C.
  \item At 12:20, the anomaly instruction expires.  Because B is again the most
  recent valid underlying route, a current query at 12:21 returns B without a
  new copy of task B being issued.
  \item At 12:25, the 09:00 high-price alert is retracted.  The retraction
  removes B from trusted history.  Consequently, a query issued at 12:30 about
  the world state at 09:15 now resolves to A, even though the same historical
  query would have resolved to B before the retraction arrived.
\end{enumerate}
At every checkpoint, the system must first reconstruct the valid route from
the evidence available at query time and then solve the selected document.
The fixed content is an experimental control: events change the route, not the
task payload, so routing and reuse can be measured without content-editing
errors.  A mutable payload would be represented as a new immutable version with
its own answer key.  The example therefore tests four distinct failure modes: allowing a
delayed event to overwrite a newer route and applying a future-effective event
too early (\emph{preservation}), failing to restore an underlying route after
expiration (\emph{propagation}), and ignoring a later retraction when answering
a historical query (\emph{historical}).

\paragraph{Collection.}
We repurpose six widely-adopted benchmarks used in Graph-of-Agents
\citep{yun2026graphofagents}: MMLU \citep{hendrycks2021mmlu}, MMLU-Pro
\citep{wang2024mmlupro}, MedMCQA
\citep{pal2022medmcqa}, MATH \citep{hendrycks2021math}, GPQA \citep{rein2024gpqa},
and HumanEval \citep{chen2021humaneval} into event streams while leaving the source
questions, answer keys, and hidden tests unchanged.  The full collection contains
31{,}119 episodes.  Each episode presents three fixed-content task documents, 12 events,
and 12 checkpoint queries: one \emph{baseline} after the first assignment, one
\emph{changed} after the first switch, four preservation, three propagation,
and three historical.  The curation protocol,
example episodes (Table~\ref{tab:episode-examples}), and collection statistics appear in
Appendix~\ref{app:dataset}.

Document selection must be scored explicitly.  Baseline and changed checkpoints
always require different documents, yet 5{,}904 of 30{,}955 exact answer episodes,
or 19.1\%, assign them the same gold answer string.  Answer only evaluation can
therefore credit a system that returns the right option while following the wrong
task.  This motivates reporting selection accuracy, answer accuracy, and their
conjunction; Appendix~\ref{app:dataset} gives the collision rates by dataset.


\section{Revision-Aware Independent Agent Graph}
\label{sec:method}

\subsection{Problem setup}
\label{sec:setup}

Let $\mathcal D=\{A,B,C\}$ index three task documents with fixed texts
$\{x_d:d\in\mathcal D\}$.  Fixed content is an experimental control: it
separates temporal binding and reuse from in-place editing, which would instead
create a new immutable version.  An episode interleaves events and queries in
arrival order.  Event $e$ has arrival time $a(e)$, effective time $\tau(e)$,
and an operation that adds, supersedes, expires, or retracts a routing
assertion.  Query $q_t$, issued at knowledge time $t$, targets world time
$w_t\le t$; equality denotes a current query and strict inequality a historical
one.  The admissible history is
\begin{equation}
    H_t = \{e : a(e) \le t\},
    \label{eq:prefix}
\end{equation}
which excludes future events, gold answers, and hidden tests.  The system returns
$(d_t,y_t)$, where $d_t$ is the document in force at $w_t$ under $H_t$ and
$y_t$ answers $x_{d_t}$.  Section~\ref{sec:experiments} scores selection,
answer, and their conjunction.

\subsection{Overview}
\label{sec:overview}
\raggedbottom

\begin{figure}[t]
  \centering
  \includegraphics[width=\textwidth]{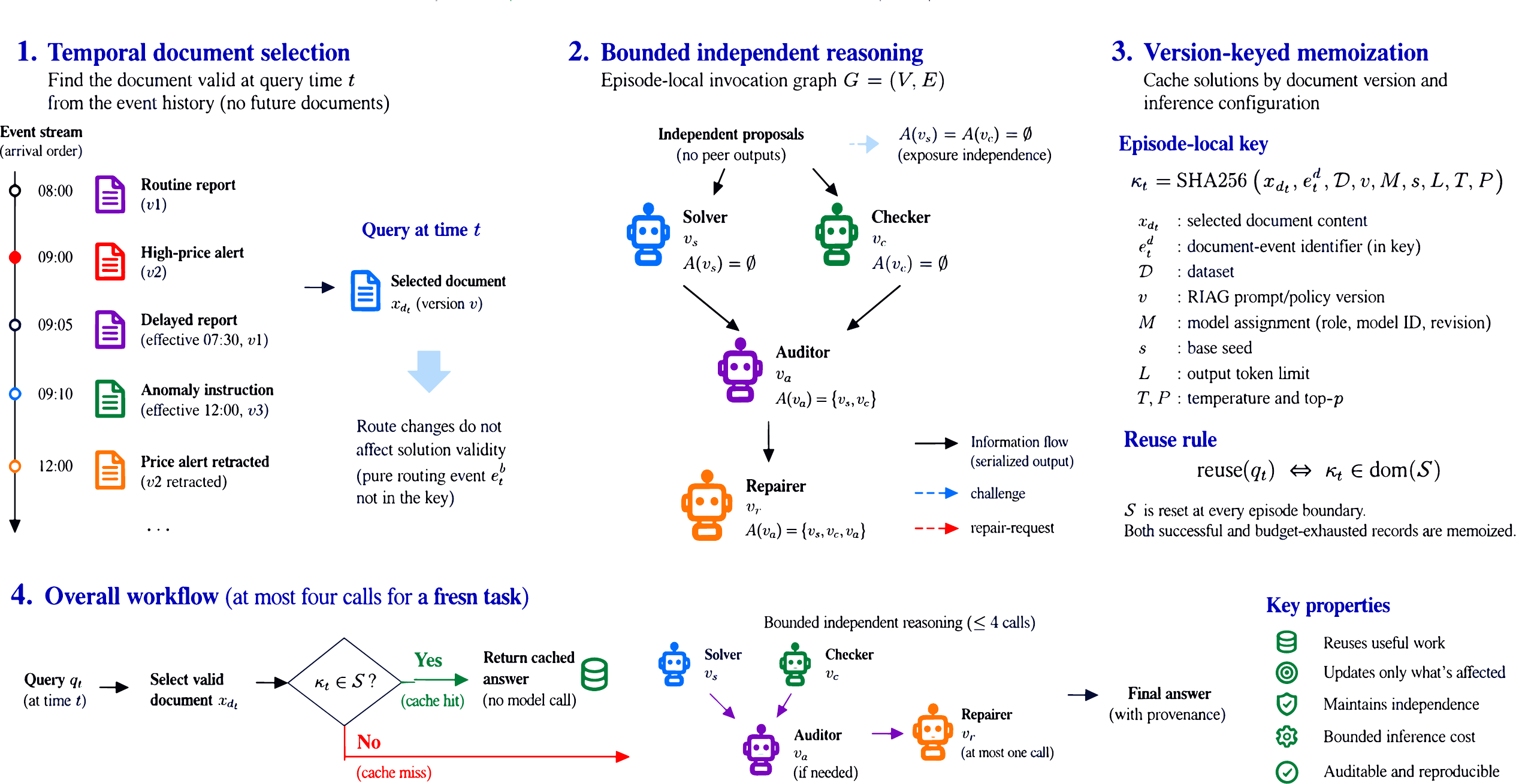}
  \caption{RIAG workflow.  A deterministic resolver selects the version valid
  at the query's world time; exact version--configuration keys enable reuse; and
  cache misses open a two-to-four-call exposure-aware graph.}
  \label{fig:riag-workflow}
\end{figure}

RIAG composes three components (Figure~\ref{fig:riag-workflow}).  First, a
deterministic resolver $\rho$ maps $(H_t,q_t)$ to the document version valid at $w_t$
without a model call (Section~\ref{sec:resolution}).  Second, an episode-local
memoizer returns a prior solution only when the document version and inference
configuration match exactly (Section~\ref{sec:memo}).  Third, a cache miss opens
a bounded graph: an unexposed solver and checker propose answers, an auditor is
called only after disagreement or invalid output, and at most one repair follows
(Section~\ref{sec:graph}).  Thus a fresh key costs two to four calls and a cache
hit costs none; every call records its exposure ancestry.

\subsection{Which task? Selecting the document in force}
\label{sec:resolution}

For each query, the resolver returns
\begin{equation}
    \rho(H_t, q_t) = \big(d_t,\; x_{d_t},\; e_t^d,\; e_t^b\big),
    \label{eq:binding}
\end{equation}
where $d_t\in\mathcal D$ is returned to the evaluator, $x_{d_t}$ is passed to
the reasoner, $e_t^d$ identifies its document version, and $e_t^b$ is the
binding assertion retained for provenance.

Let $\mathcal A$ be the routing-assertion events.  For assertion $e$, define
$\operatorname{closed}_t(e,w)$ to hold when $H_t$ contains a supersession or
expiration $s$ targeting $e$ with $\tau(s)\le w$, and define
$\operatorname{retracted}_t(e)$ to hold when $H_t$ contains a retraction $r$
targeting $e$ with $\tau(r)\le t$.  The assertions valid for world time $w$ are
\begin{equation}
 \mathcal V_t(w)=\{e\in\mathcal A:a(e)\le t,\ \tau(e)\le w,
 \ \neg\operatorname{closed}_t(e,w),\
 \ \neg\operatorname{retracted}_t(e)\}.
 \label{eq:valid-assertions}
\end{equation}
For a delivery request, $e_t^b$ is the unique maximizer in $\mathcal V_t(w_t)$
under the lexicographic order $(\tau(e),a(e),\operatorname{id}(e))$, and
$d_t=\operatorname{doc}(e_t^b)$.  For an archive request, $e_t^b$ is its fixed
binding if that binding remains valid.  The active document-version assertion
for $d_t$ is $e_t^d$; malformed or non-unique states fail resolution.  This
separates world-time validity from knowledge-time retraction: once a retraction
is known and effective, it revises later answers even about earlier world times.
The resolver recomputes Equation~\ref{eq:binding} at every query and makes no
model call; Appendix~\ref{app:resolver} gives the executable edge cases.

\subsection{Recompute or reuse? Version-keyed memoization}
\label{sec:memo}

A stored solution is reusable only for the same document version and inference
configuration.  Let $\operatorname{enc}$ be the implementation's canonical,
length-delimited serialization.  RIAG defines
\begin{equation}
    \kappa_t = \operatorname{SHA256}\!\left(\operatorname{enc}
    (x_{d_t},e_t^d,\mathcal D,v,M,s,L,T,P)\right),
    \label{eq:fingerprint}
\end{equation}
where $\mathcal D$ is the dataset; $v$ the prompt and policy version; $M$ the
ordered role-to-model assignment; $s$ the base seed; $L$ the output limit; and
$(T,P)$ the temperature and top-$p$.  The pair $(x_{d_t},e_t^d)$ identifies the
selected version.  With episode-local solution table $\mathcal S$, the reuse
rule is
\begin{equation}
    \operatorname{reuse}(q_t) \iff \kappa_t \in \operatorname{dom}(\mathcal{S}).
    \label{eq:reuse}
\end{equation}
The binding event $e_t^b$ is deliberately absent: leaving and later returning
to the same version reuses its solution.  By contrast, a new $e_t^d$ or any
change to $M,v,s,L,T,$ or $P$ forces recomputation, even if the text or semantics
appear unchanged.  This conservative rule assumes collision-free SHA-256 on
the realized keys and uses only public identifiers; it does not infer
proposition-level dependencies.  $\mathcal S$ is cleared at each episode start,
and failed or budget-exhausted records are also cached so that repetition cannot
reset the call budget.  Appendix~\ref{app:resolver} specifies serialization.

\subsection{Whose agreement counts? A bounded, exposure-aware graph}
\label{sec:graph}
\label{sec:policy}

On a cache miss, RIAG constructs a directed acyclic invocation graph
$G=(V,E)$ with at most four model-call nodes.  Edge $(u,v)\in E$ means that
$u$'s output appears in $v$'s prompt.  Exposure ancestry is defined recursively
by
\begin{equation}
 A(v)=\{u:(u,v)\in E\}\ \cup\!
 \bigcup_{(u,v)\in E} A(u).
 \label{eq:exposure-ancestry}
\end{equation}
Slots are reserved before dispatch; a timeout or malformed reply consumes its
slot and triggers no hidden retry.  Figure~\ref{fig:riag-workflow} shows the
three possible stages.

\paragraph{Independent proposals.}
Solver $v_{\mathrm{s}}$ and checker $v_{\mathrm{c}}$ receive the task and
role-specific instructions but no peer output, hence
\begin{equation}
    A(v_{\mathrm{s}}) = A(v_{\mathrm{c}}) = \varnothing.
    \label{eq:exposure-independence}
\end{equation}
Both commit before either output is disclosed, including under parallel
execution.  This \emph{exposure independence} prevents agreement caused by
peer observation but does not imply statistically independent errors.  If both
parsed answers are valid and equal, RIAG returns the solver's answer after two
calls.

\paragraph{Targeted audit.}
If either proposal is invalid or the answers disagree, auditor
$v_{\mathrm{a}}$ receives both records, so
$A(v_{\mathrm{a}})=\{v_{\mathrm{s}},v_{\mathrm{c}}\}$.  It may select a
candidate or recompute disputed steps.  A valid, justified output that does not
request repair terminates after three calls.

\paragraph{Bounded repair.}
Otherwise repairer $v_{\mathrm{r}}$ receives all three records, with
$A(v_{\mathrm{r}})=\{v_{\mathrm{s}},v_{\mathrm{c}},v_{\mathrm{a}}\}$.  A
valid, justified repair is returned after four calls but marked unresolved
because no later call verifies it.  If repair fails, RIAG returns the first
valid initial answer (solver before checker), or $\bot$ if neither is valid.

Formally, let $\operatorname{valid}(v)$ mean that $v$ passes the public
dataset-specific format check; let $\operatorname{usable}(v)$ additionally
require a nonempty justification; and let $\operatorname{repair}(v_a)$ denote
an audit request for repair.  Evaluating branches from top to bottom, RIAG emits
\begin{equation}
 \widehat y_t=
 \begin{cases}
 y_{\mathrm{s}}, & \operatorname{valid}(v_{\mathrm{s}})\wedge
                    \operatorname{valid}(v_{\mathrm{c}})\wedge
                    y_{\mathrm{s}}=y_{\mathrm{c}},\\
 y_{\mathrm{a}}, & \operatorname{usable}(v_{\mathrm{a}})\wedge
                    \neg\operatorname{repair}(v_{\mathrm{a}}),\\
 y_{\mathrm{r}}, & \operatorname{usable}(v_{\mathrm{r}}),\\
 y_{\mathrm{s}}, & \operatorname{valid}(v_{\mathrm{s}}),\\
 y_{\mathrm{c}}, & \operatorname{valid}(v_{\mathrm{c}}),\\
 \bot, & \text{otherwise},
 \end{cases}
 \label{eq:decision}
\end{equation}
The public checks accept a listed option letter, a nonempty
\verb|\boxed{}| expression, or Python defining the required entry point; they
do not test semantic correctness.  Thus solver--checker agreement is evidence
from two unexposed attempts, whereas auditor and repairer outputs are
adjudications of exposed evidence, not additional independent votes.
Appendix~\ref{app:resolver} specifies the checks and invocation records.

\subsection{Instantiations}
\label{sec:instantiations}

Following GoA \citep{yun2026graphofagents}, we evaluate two role-to-model maps
under the same resolver, memoizer, prompts, parsers, branching rule, and call
cap.  \emph{Homogeneous} RIAG assigns Qwen2.5-7B-Instruct
\citep{yang2024qwen25} to all roles.  \emph{Heterogeneous} RIAG assigns
Qwen2.5-7B-Instruct to the solver, Mathstral-7B-v0.1
\citep{mistralai2024mathstral} to the checker, Mistral-7B-Instruct-v0.3
\citep{jiang2023mistral7b} to the auditor, and Qwen2.5-Coder-7B-Instruct
\citep{hui2024qwen25coder} to the repairer.  The complete ordered map is part of
$\kappa_t$, so changing any role's model invalidates reuse.

\section{Experiments}
\label{sec:experiments}

We evaluate RIAG on all 31,119 episodes and 373,428 scheduled queries across six
dynamic task routing datasets.  Comparisons include Debate, Self-Consistency,
Refine, ReConcile, Mixture-of-Agents, Self-MoA, an answer-generation adaptation
of CIA, and Graph-of-Agents (GoA) with max or mean pooling and homogeneous or
heterogeneous model pools.  Each full-collection run uses seed~0.
Appendix~\ref{app:detailed-experiments} provides the full protocol and results.

\subsection{Overall results}

Table~\ref{tab:overall} reports four metrics.  Selection accuracy is the fraction
of queries routed to the gold document; answer accuracy is the fraction answered
correctly for the gold task; and joint accuracy requires both on the same query.
Calls/q averages recorded model invocations over scheduled queries, counting a
cache hit as zero calls.

The five category columns partition the queries.  Baseline tests the initial
assignment; changed, a direct update; propagation, an update that should alter
the route; preservation, an event that should not alter it; and historical, the
route at an earlier world time.  Each column reports joint accuracy within that
category.  Overall Joint is their query-weighted aggregate, not a sixth category
or a simple mean.  With one baseline, one changed, three propagation, four
preservation, and three historical queries per episode, it is
$\left(A_B+A_C+3A_{Pr}+4A_{Pe}+3A_H\right)/12$.

Debate is the strongest comparison method, with 32.22\% joint accuracy at 18.00
calls/q.  Homogeneous RIAG reaches 54.24\% with perfect selection at 0.62
calls/q; heterogeneous RIAG reaches 49.78\% with 98.81\% selection at 0.63
calls/q.  Homogeneous RIAG is stable across categories (53.89--54.50\%), while
Debate falls from 53.06\% on baseline to 26.48\% on historical queries.
Preservation carries the largest weight in Overall Joint, making performance on
unchanged routes especially important.

The comparison methods select the correct document on roughly 60\% of queries,
and their answer accuracy exceeds joint accuracy by about seven points.  Their
errors therefore reflect both routing and task solution; answer accuracy alone
can credit an answer produced from the wrong document.  RIAG also uses about
29 times fewer calls than Debate and 42 times fewer than adapted CIA, primarily
because it reuses prior solutions.

\begin{table}[htbp]
\centering
\scriptsize
\setlength{\tabcolsep}{2pt}
\renewcommand{\arraystretch}{0.95}
\setlength{\abovecaptionskip}{3pt}
\setlength{\belowcaptionskip}{2pt}
\caption{Performance on the dynamic task routing variants of MMLU, MMLU-Pro, MedMCQA, MATH, GPQA, and HumanEval. Dataset-wise breakdowns are provided in Appendix \ref{appedix:perf_breakdown}.}\label{tab:overall}
\begin{tabularx}{\linewidth}{@{}ll*{9}{>{\centering\arraybackslash}X}@{}}
\toprule
 &  & \multicolumn{3}{c}{Accuracy (\%) $\uparrow$} & Cost $\downarrow$ & \multicolumn{5}{c}{Joint accuracy (\%) by query category $\uparrow$}\\
\cmidrule(lr){3-5}\cmidrule(lr){6-6}\cmidrule(l){7-11}
 &  & \textbf{Joint} & \textbf{Answer} & \textbf{Selection} & \textbf{Calls/q} & \textbf{Baseline} & \textbf{Changed} & \textbf{Propag.} & \textbf{Preserv.} & \textbf{Historical}\\
\midrule
 & Debate & 32.22 & 39.42 & 62.78 & 18.00 & \underline{53.06} & \underline{50.94} & 34.71 & 24.78 & 26.48\\
 & Self-Consistency & 26.62 & 34.33 & 61.40 & 6.00 & 45.70 & 37.14 & 27.56 & 22.50 & 21.29\\
 & Refine & 27.84 & 35.34 & 60.85 & 19.00 & 48.49 & 42.94 & 29.96 & 20.75 & 23.25\\
 & ReConcile & 28.92 & 35.37 & 63.51 & 18.00 & 49.42 & 43.83 & 31.08 & 21.98 & 24.22\\
 & Mixture-of-Agents & 28.98 & 36.37 & 61.06 & 19.00 & 50.30 & 47.25 & 32.15 & 20.88 & 23.42\\
 & Self-MoA & 26.17 & 33.78 & 60.95 & 7.00 & 45.48 & 40.04 & 27.87 & 20.42 & 21.08\\
\multirow{-7}{*}{\textbf{Baselines}} & CIA (adapted) & 29.36 & 36.90 & 61.70 & 26.00 & 51.26 & 40.84 & 32.49 & 23.22 & 23.27  \\
\midrule
 & max, homogeneous & 28.18 & 34.99 & 60.58 & 10.58 & 49.60 & 43.16 & 30.51 & 21.09 & 23.16\\
 & mean, homogeneous & 28.94 & 36.47 & 59.17 & 12.56 & 51.60 & 46.07 & 30.99 & 21.61 & 23.40\\
 & max, heterogeneous & 28.65 & 35.94 & 60.23 & 11.45 & 49.66 & 45.71 & 31.13 & 21.37 & 23.20\\
\multirow{-4}{*}{\shortstack[l]{\textbf{Graph-of-}\\\textbf{Agents}}} & mean, heterogeneous & 28.68 & 36.33 & 58.68 & 13.48 & 50.13 & 45.72 & 30.65 & 21.50 & 23.45\\
\midrule
\rowcolor{gray!15} & homogeneous & \textbf{54.24} & \textbf{54.24} & \textbf{100.00} & \textbf{0.62} & \textbf{53.89} & \textbf{54.25} & \textbf{54.50} & \textbf{54.10} & \textbf{54.29}\\
\rowcolor{gray!15}\multirow{-2}{*}{\shortstack[l]{\textbf{RIAG}\\\textit{(ours)}}} & heterogeneous & \underline{49.78} & \underline{49.78} & \underline{98.81} & \underline{0.63} & 50.05 & 49.31 & \underline{49.84} & \underline{49.77} & \underline{49.81}\\
\bottomrule
\end{tabularx}
\end{table}

\subsection{Temporal behavior, controls, and robustness}

The advantage is consistent across datasets and temporal conditions.
Homogeneous RIAG obtains 40.10--66.83\% joint accuracy across the six datasets,
versus 23.02--38.66\% for the best comparison method on each dataset.  It also
remains within 53.89--54.50\% across the five query categories, whereas every
conventional method loses 24--29 points from baseline to historical queries.
Table~\ref{tab:overall} and Appendix Table~\ref{tab:joint-dataset} give both
breakdowns.

\begin{figure}[t]
  \centering
  \includegraphics[width=.9\linewidth]{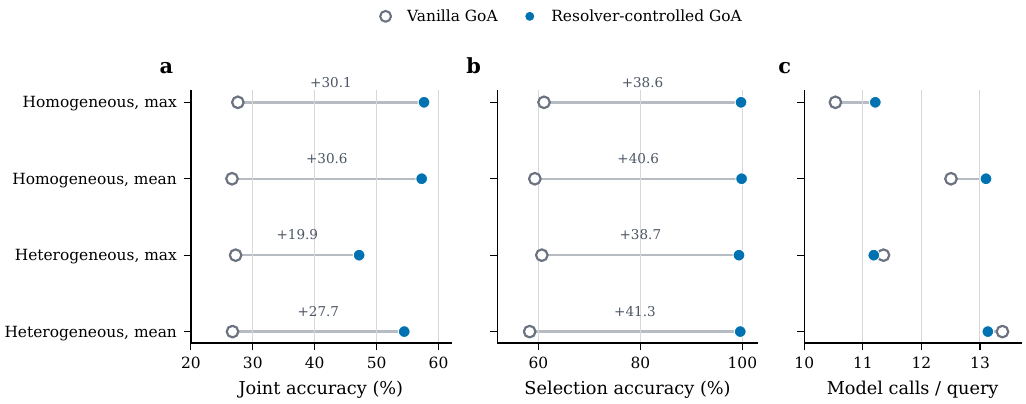}
  \caption{Analysis of the transferability of the proposed deterministic resolver design to Graph-of-Agents on MATH (500 episodes and 6,000 queries per configuration). Lines pair vanilla and
  resolver-controlled runs with the same pooling and model-pool labels.
  (a) Joint accuracy, with changes in percentage points; (b) selection accuracy;
  (c) recorded calls/query.}
  \label{fig:goa-math-resolver-control}
\end{figure}

To test whether resolver control is specific to RIAG, we place the same
deterministic resolver in front of GoA while retaining GoA's max/mean pooling
protocol and omitting RIAG's cache and bounded graph.  Figure
\ref{fig:goa-math-resolver-control} shows gains for all four model-pool and
pooling combinations on MATH.  Selection rises from 58.30--61.12\% for vanilla
GoA to 99.35--99.87\%, while joint accuracy improves by 30.07 and 30.63 points
for homogeneous max and mean pooling and by 19.95 and 27.73 points for their
heterogeneous counterparts.  The common improvement across these GoA variants
shows that deterministic temporal resolution can serve as a reusable front end
for a different reasoning-graph family, rather than only as an internal RIAG
component.

After resolution, max and mean differ by only 0.38 points under homogeneous
assignment, while mean leads max by 7.28 points under heterogeneous assignment.
Parse failures fall from 369 to 86 across the four runs.  The controlled package
also changes role cards and decoding temperature, so these results establish
architectural compatibility and recorded end-to-end gains, not the resolver's
isolated causal effect.  Appendix Section~\ref{sec:full-math-analysis} provides
the full analysis.

A separate RIAG sensitivity sweep yields 62.05--64.50\% joint accuracy for the
homogeneous pool and 49.70--57.02\% for the heterogeneous pool.  Only two
changes have paired intervals excluding zero: a 1,024-token homogeneous limit
($-2.08$ points) and heterogeneous temperature 1.0 ($-5.18$ points).  The former
causes 115 format failures; the latter causes none, indicating different
truncation and sampling effects.


\subsection{Reuse, conditional expansion, and response diversity}

The reference MATH runs are cache hits on 76.9\% of queries.  On fresh tasks,
homogeneous RIAG stops after two agreeing calls in 66.2\% of cases, invokes the
auditor in 30.2\%, and reaches repair in 3.5\%.  The corresponding heterogeneous
rates are 41.5\%, 51.3\%, and 7.2\%.  This deeper expansion costs more without
higher accuracy: homogeneous RIAG reaches 64.13\% at 0.548 calls/q, versus
54.88\% at 0.614 calls/q for heterogeneous RIAG.

Reuse follows the event semantics: baseline and changed queries always open a
fresh graph, 74.3\% of propagation queries reuse a solution, and all preservation
and historical queries do so.  On fresh tasks, the two-call agreement route is
accurate in 82.1\% of homogeneous and 85.4\% of heterogeneous cases.  Audited
and repaired cases are harder, but their conditional accuracy does not measure
the value of extra calls because disagreement or failed validation triggers
those routes.

\begin{figure}[t]
  \centering
  \begin{minipage}[t]{0.455\linewidth}
    \centering
    \includegraphics[width=\linewidth]{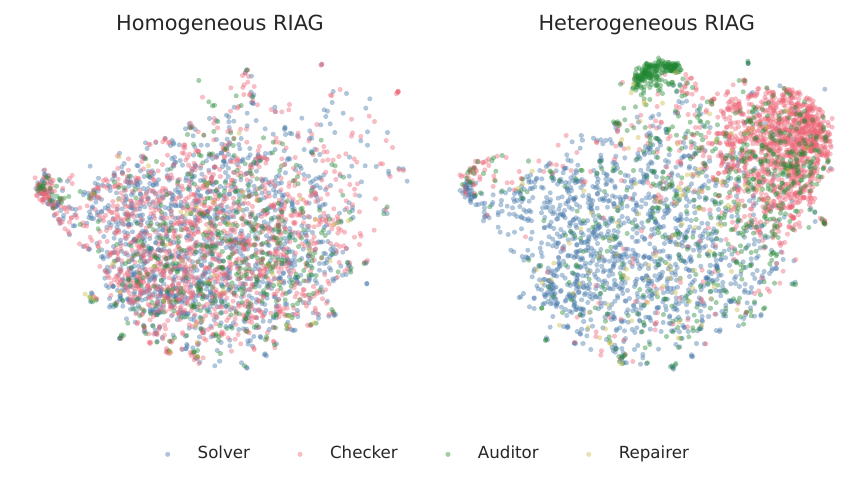}
  \end{minipage}\hfill
  \begin{minipage}[t]{0.535\linewidth}
    \centering
    \includegraphics[width=\linewidth]{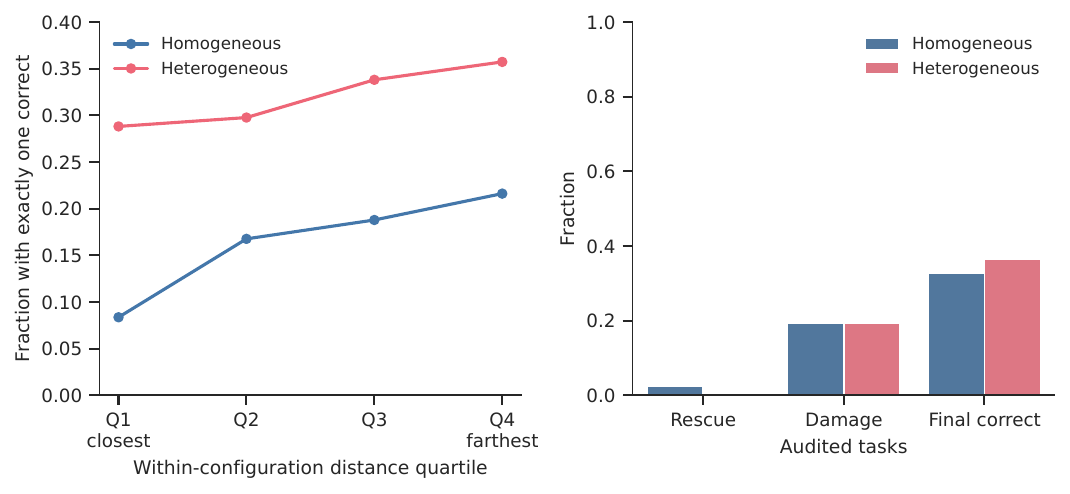}
  \end{minipage}
  \caption{Reasoning geometry and conditional intervention.  Left: task-centered
  response UMAP for RIAG invocations, jointly fitted across homogeneous and
  heterogeneous roles.  Right: useful diversity by solver--checker distance
  quartile and outcomes conditional on auditing.  Rescue means that two wrong
  initial answers become a correct final answer; damage means that at least one
  correct initial answer becomes a wrong final answer.}
  \label{fig:riag-role-umap}
  \label{fig:riag-distance-audit}
\end{figure}

Figure~\ref{fig:riag-role-umap} asks whether greater response diversity is
useful.  A role probe reaches 34.3\% balanced accuracy for homogeneous RIAG
(25\% chance) and 65.7\% for heterogeneous RIAG, where role and model are
confounded.  Heterogeneity increases mean solver--checker cosine distance from
0.111 to 0.190, a paired increase of 0.080 (95\% task-bootstrap interval
$[0.075,0.084]$), and produces the larger distance on 83.8\% of matched tasks.

That separation does not improve outcomes.  Useful diversity rises from 16.4\%
to 32.0\%, but initial answer coverage falls from 73.4\% to 68.8\% and final
unique-task accuracy from 67.6\% to 57.5\%.  From the closest to farthest
distance quartile, coverage also falls from 80.7\% to 62.8\% for homogeneous
RIAG and from 79.8\% to 55.9\% for heterogeneous RIAG.  Among the 478 homogeneous
and 830 heterogeneous tasks selected for auditing, rescue rates are 2.5\% and
0.6\%, damage rates are 19.2\% and 19.4\%, and final accuracies are 32.8\% and
36.5\%.  These rates are descriptive because difficult cases trigger auditing.
Appendix Section~\ref{sec:math-reasoning-umap} provides the full analysis.

%

\section{Conclusion}
\label{sec:conclusion}

We introduce RIAG, a bounded inference policy that separates temporal task resolution
from task reasoning, so that solutions are reused across event revisions and agreement
is counted only when it precedes communication.  Experiments on six dynamic task
routing datasets show higher joint accuracy at fewer than one model call per query,
and the resolver-controlled GoA study shows that the temporal front end composes
with a different reasoning-graph family.  We hope this separation helps future
work and agents that operate over evolving event streams.

\bibliographystyle{iclr2027_conference}
\bibliography{iclr2027_conference}

@inproceedings{yun2026graphofagents,
title={Graph-of-Agents: A Graph-based Framework for Multi-Agent {LLM} Collaboration},
author={Sukwon Yun and Jie Peng and Pingzhi Li and Wendong Fan and Jie Chen and James Zou and Guohao Li and Tianlong Chen},
booktitle={The Fourteenth International Conference on Learning Representations},
year={2026}
}

@inproceedings{hendrycks2021mmlu,
title={Measuring Massive Multitask Language Understanding},
author={Hendrycks, Dan and Burns, Collin and Basart, Steven and Zou, Andy and Mazeika, Mantas and Song, Dawn and Steinhardt, Jacob},
booktitle={International Conference on Learning Representations},
year={2021},
url={https://openreview.net/forum?id=d7KBjmI3GmQ}
}

@inproceedings{wang2024mmlupro,
title={{MMLU-Pro}: A More Robust and Challenging Multi-Task Language Understanding Benchmark},
author={Wang, Yubo and Ma, Xueguang and Zhang, Ge and Ni, Yuansheng and Chandra, Abhranil and Guo, Shiguang and Ren, Weiming and Arulraj, Aaran and He, Xuan and Jiang, Ziyan and Li, Tianle and Ku, Max and Wang, Kai and Zhuang, Alex and Fan, Rongqi and Yue, Xiang and Chen, Wenhu},
booktitle={Advances in Neural Information Processing Systems},
volume={37},
year={2024},
doi={10.52202/079017-3018}
}

@inproceedings{pal2022medmcqa,
title={{MedMCQA}: A Large-scale Multi-Subject Multi-Choice Dataset for Medical Domain Question Answering},
author={Pal, Ankit and Umapathi, Logesh Kumar and Sankarasubbu, Malaikannan},
booktitle={Proceedings of the Conference on Health, Inference, and Learning},
series={Proceedings of Machine Learning Research},
volume={174},
pages={248--260},
publisher={PMLR},
year={2022},
url={https://proceedings.mlr.press/v174/pal22a.html}
}

@inproceedings{hendrycks2021math,
title={Measuring Mathematical Problem Solving With the {MATH} Dataset},
author={Hendrycks, Dan and Burns, Collin and Kadavath, Saurav and Arora, Akul and Basart, Steven and Tang, Eric and Song, Dawn and Steinhardt, Jacob},
booktitle={Proceedings of the Neural Information Processing Systems Track on Datasets and Benchmarks},
volume={1},
year={2021}
}

@inproceedings{rein2024gpqa,
title={{GPQA}: A Graduate-Level Google-Proof Q\&A Benchmark},
author={Rein, David and Hou, Betty Li and Stickland, Asa Cooper and Petty, Jackson and Pang, Richard Yuanzhe and Dirani, Julien and Michael, Julian and Bowman, Samuel R.},
booktitle={Conference on Language Modeling},
year={2024},
url={https://openreview.net/forum?id=Ti67584b98}
}

@article{chen2021humaneval,
title={Evaluating Large Language Models Trained on Code},
author={Chen, Mark and others},
journal={arXiv preprint arXiv:2107.03374},
year={2021},
doi={10.48550/arXiv.2107.03374}
}

@article{yang2024qwen25,
title={{Qwen2.5} Technical Report},
author={Yang, An and others},
journal={arXiv preprint arXiv:2412.15115},
year={2024},
doi={10.48550/arXiv.2412.15115}
}

@misc{mistralai2024mathstral,
title={{Mathstral}},
author={{Mistral AI}},
year={2024},
howpublished={Mistral AI Research Release},
url={https://mistral.ai/news/mathstral/}
}

@article{jiang2023mistral7b,
title={Mistral 7B},
author={Jiang, Albert Q. and Sablayrolles, Alexandre and Mensch, Arthur and Bamford, Chris and Chaplot, Devendra Singh and de las Casas, Diego and Bressand, Florian and Lengyel, Gianna and Lample, Guillaume and Saulnier, Lucile and Lavaud, L{\'e}lio Renard and Lachaux, Marie-Anne and Stock, Pierre and Le Scao, Teven and Lavril, Thibaut and Wang, Thomas and Lacroix, Timoth{\'e}e and El Sayed, William},
journal={arXiv preprint arXiv:2310.06825},
year={2023},
doi={10.48550/arXiv.2310.06825}
}

@article{hui2024qwen25coder,
title={{Qwen2.5-Coder} Technical Report},
author={Hui, Binyuan and others},
journal={arXiv preprint arXiv:2409.12186},
year={2024},
doi={10.48550/arXiv.2409.12186}
}

@article{zhuge2024gptswarm,
title={Language Agents as Optimizable Graphs},
author={Zhuge, Mingchen and Wang, Wenyi and Kirsch, Louis and Faccio, Francesco and Khizbullin, Dmitrii and Schmidhuber, J{\"u}rgen},
journal={arXiv preprint arXiv:2402.16823},
year={2024},
doi={10.48550/arXiv.2402.16823}
}

@inproceedings{zhang2025aflow,
title={{AFlow}: Automating Agentic Workflow Generation},
author={Zhang, Jiayi and Xiang, Jinyu and Yu, Zhaoyang and Teng, Fengwei and Chen, Xiong-Hui and Chen, Jiaqi and Zhuge, Mingchen and Cheng, Xin and Hong, Sirui and Wang, Jinlin and Zheng, Bingnan and Liu, Bang and Luo, Yuyu and Wu, Chenglin},
booktitle={International Conference on Learning Representations},
year={2025},
url={https://openreview.net/forum?id=z5uVAKwmjf}
}

@article{du2023multiagentdebate,
title={Improving Factuality and Reasoning in Language Models through Multiagent Debate},
author={Du, Yilun and Li, Shuang and Torralba, Antonio and Tenenbaum, Joshua B. and Mordatch, Igor},
journal={arXiv preprint arXiv:2305.14325},
year={2023},
doi={10.48550/arXiv.2305.14325}
}

@inproceedings{chen2024reconcile,
title={{ReConcile}: Round-Table Conference Improves Reasoning via Consensus among Diverse {LLM}s},
author={Chen, Justin Chih-Yao and Saha, Swarnadeep and Bansal, Mohit},
booktitle={Proceedings of the 62nd Annual Meeting of the Association for Computational Linguistics},
pages={7066--7085},
publisher={Association for Computational Linguistics},
year={2024},
doi={10.18653/v1/2024.acl-long.381}
}

@article{wang2024mixtureofagents,
title={Mixture-of-Agents Enhances Large Language Model Capabilities},
author={Wang, Junlin and Wang, Jue and Athiwaratkun, Ben and Zhang, Ce and Zou, James},
journal={arXiv preprint arXiv:2406.04692},
year={2024},
doi={10.48550/arXiv.2406.04692}
}

@article{wu2023autogen,
title={{AutoGen}: Enabling Next-Gen {LLM} Applications via Multi-Agent Conversation},
author={Wu, Qingyun and Bansal, Gagan and Zhang, Jieyu and Wu, Yiran and Li, Beibin and Zhu, Erkang and Jiang, Li and Zhang, Xiaoyun and Zhang, Shaokun and Liu, Jiale and Awadallah, Ahmed Hassan and White, Ryen W. and Burger, Doug and Wang, Chi},
journal={arXiv preprint arXiv:2308.08155},
year={2023},
doi={10.48550/arXiv.2308.08155}
}

@inproceedings{park2023generativeagents,
title={Generative Agents: Interactive Simulacra of Human Behavior},
author={Park, Joon Sung and O'Brien, Joseph C. and Cai, Carrie J. and Morris, Meredith Ringel and Liang, Percy and Bernstein, Michael S.},
booktitle={Proceedings of the 36th Annual ACM Symposium on User Interface Software and Technology},
pages={1--22},
publisher={ACM},
year={2023},
doi={10.1145/3586183.3606763}
}

@inproceedings{shinn2023reflexion,
title={Reflexion: Language Agents with Verbal Reinforcement Learning},
author={Shinn, Noah and Cassano, Federico and Gopinath, Ashwin and Narasimhan, Karthik and Yao, Shunyu},
booktitle={Advances in Neural Information Processing Systems},
volume={36},
year={2023}
}

@article{packer2023memgpt,
title={{MemGPT}: Towards {LLM}s as Operating Systems},
author={Packer, Charles and Wooders, Sarah and Lin, Kevin and Fang, Vivian and Patil, Shishir G. and Stoica, Ion and Gonzalez, Joseph E.},
journal={arXiv preprint arXiv:2310.08560},
year={2023},
doi={10.48550/arXiv.2310.08560}
}

@inproceedings{wang2023selfconsistency,
  title={Self-Consistency Improves Chain of Thought Reasoning in Language Models},
  author={Wang, Xuezhi and Wei, Jason and Schuurmans, Dale and Le, Quoc V. and
          Chi, Ed H. and Narang, Sharan and Chowdhery, Aakanksha and Zhou, Denny},
  booktitle={International Conference on Learning Representations},
  year={2023},
  doi={10.48550/arXiv.2203.11171}
}

@inproceedings{madaan2023selfrefine,
  title={Self-Refine: Iterative Refinement with Self-Feedback},
  author={Madaan, Aman and Tandon, Niket and Gupta, Prakhar and Hallinan, Skyler and
          Gao, Luyu and Wiegreffe, Sarah and Alon, Uri and Dziri, Nouha and
          Prabhumoye, Shrimai and Yang, Yiming and Gupta, Shashank and
          Majumder, Bodhisattwa Prasad and Hermann, Katherine and Welleck, Sean and
          Yazdanbakhsh, Amir and Clark, Peter},
  booktitle={Advances in Neural Information Processing Systems},
  volume={36},
  year={2023},
  doi={10.48550/arXiv.2303.17651}
}

@article{li2025selfmoa,
  title={Rethinking Mixture-of-Agents: Is Mixing Different Large Language Models Beneficial?},
  author={Li, Wenzhe and Lin, Yong and Xia, Mengzhou and Jin, Chi},
  journal={arXiv preprint arXiv:2502.00674},
  year={2025},
  doi={10.48550/arXiv.2502.00674}
}

@inproceedings{wu2026cia,
  title={{CIA}: Inferring the Communication Topology from {LLM}-based Multi-Agent Systems},
  author={Wu, Yongxuan and Lin, Xixun and Zhang, He and Sun, Nan and Wang, Kun and
          Zhou, Chuan and Pan, Shirui and Cao, Yanan},
  booktitle={Proceedings of the 64th Annual Meeting of the Association for
             Computational Linguistics (Volume 1: Long Papers)},
  pages={17917--17933},
  year={2026},
  url={https://aclanthology.org/2026.acl-long.815/}
}

@inproceedings{smit2024mad,
  title={Should We Be Going {MAD}? A Look at Multi-Agent Debate Strategies for {LLM}s},
  author={Smit, Andries Petrus and Grinsztajn, Nathan and Duckworth, Paul and
          Barrett, Thomas D and Pretorius, Arnu},
  booktitle={Proceedings of the 41st International Conference on Machine Learning},
  volume={235},
  year={2024},
  url={https://proceedings.mlr.press/v235/smit24a.html}
}

@inproceedings{reimers2019sentencebert,
  title={Sentence-{BERT}: Sentence Embeddings using {S}iamese {BERT}-Networks},
  author={Reimers, Nils and Gurevych, Iryna},
  booktitle={Proceedings of the 2019 Conference on Empirical Methods in Natural Language
             Processing and the 9th International Joint Conference on Natural Language
             Processing (EMNLP-IJCNLP)},
  pages={3982--3992},
  publisher={Association for Computational Linguistics},
  year={2019},
  doi={10.18653/v1/D19-1410}
}

@article{mcinnes2018umap,
  title={{UMAP}: Uniform Manifold Approximation and Projection for Dimension Reduction},
  author={McInnes, Leland and Healy, John and Melville, James},
  journal={arXiv preprint arXiv:1802.03426},
  year={2018},
  doi={10.48550/arXiv.1802.03426}
}

@article{wang2024voyager,
  title={Voyager: An Open-Ended Embodied Agent with Large Language Models},
  author={Wang, Guanzhi and Xie, Yuqi and Jiang, Yunfan and Mandlekar, Ajay and
          Xiao, Chaowei and Zhu, Yuke and Fan, Linxi and Anandkumar, Anima},
  journal={Transactions on Machine Learning Research},
  year={2024},
  url={https://openreview.net/forum?id=ehfRiF0R3a}
}

@inproceedings{yang2024sweagent,
  title={{SWE-agent}: Agent-Computer Interfaces Enable Automated Software Engineering},
  author={Yang, John and Jimenez, Carlos E. and Wettig, Alexander and Lieret, Kilian and
          Yao, Shunyu and Narasimhan, Karthik and Press, Ofir},
  booktitle={Advances in Neural Information Processing Systems},
  volume={37},
  year={2024},
  doi={10.48550/arXiv.2405.15793}
}

@inproceedings{liu2024agentbench,
  title={{AgentBench}: Evaluating {LLM}s as Agents},
  author={Liu, Xiao and Yu, Hao and Zhang, Hanchen and Xu, Yifan and Lei, Xuanyu and
          Lai, Hanyu and Gu, Yu and Ding, Hangliang and Men, Kaiwen and Yang, Kejuan and
          Zhang, Shudan and Deng, Xiang and Zeng, Aohan and Du, Zhengxiao and
          Zhang, Chenhui and Shen, Sheng and Zhang, Tianjun and Su, Yu and Sun, Huan and
          Huang, Minlie and Dong, Yuxiao and Tang, Jie},
  booktitle={International Conference on Learning Representations},
  year={2024},
  doi={10.48550/arXiv.2308.03688}
}

@inproceedings{ma2024agentboard,
  title={{AgentBoard}: An Analytical Evaluation Board of Multi-turn {LLM} Agents},
  author={Ma, Chang and Zhang, Junlei and Zhu, Zhihao and Yang, Cheng and Yang, Yujiu and
          Jin, Yaohui and Lan, Zhenzhong and Kong, Lingpeng and He, Junxian},
  booktitle={Advances in Neural Information Processing Systems, Datasets and Benchmarks Track},
  volume={37},
  year={2024},
  doi={10.48550/arXiv.2401.13178}
}

\appendix
\vspace{2.5\baselineskip}
\section*{Appendix}
\makeatletter
\let\riag@origsection\section
\renewcommand{\section}{\vspace{0.8\baselineskip}\riag@origsection}
\makeatother
\section{Use of Large Language Models}
\label{app:ai-use}

We used GPT-6 Astra, a large language model, for grammar checking and refining the
text of this paper.  Astra was also used during dataset curation to help develop
and inspect the conversion of the six source benchmarks into dynamic task
routing episodes.  It did not change any source question,
option, answer key, code entry point, or hidden test.  The released episodes are
generated by a deterministic builder from the frozen source files and were
checked by three human reviewers and an independent validator
(Appendix~\ref{sec:dataset-curation}).  The authors take full responsibility
for the paper's contents.

%

\section{Reproducibility Statement}
\label{app:reproducibility}

We provide the codebase and the dynamic task routing collection in the
supplementary materials.  The collection is built from six publicly available
benchmarks by a seeded, deterministic builder; its curation protocol and
statistics are given in Appendix~\ref{app:dataset}.  Model checkpoints, decoding
settings, and context lengths for every method are listed in
Appendix~\ref{app:detailed-experiments}.  All experiments are conducted on a
workstation running Ubuntu 22.04 LTS with two NVIDIA H100 GPUs.

\section{Ethics Statement}
\label{app:ethics}

All data come from six public benchmarks, and the synthetic events we add only
change which task is active; no personal or human-subject data is used.  RIAG
inherits the biases and factual errors of the models it runs.  Its answers on
medical questions are benchmark results and should not guide clinical
decisions.

\section{Detailed Experimental Results}
\label{app:detailed-experiments}

\subsection{Evaluation protocol}

We evaluate every method on the full collection from
Section~\ref{sec:routing-data}: 31,119 episodes and 373,428 scheduled queries
across GPQA (2,376), MATH (6,000), HumanEval (1,968), MedMCQA (50,196),
MMLU-Pro (144,384), and MMLU (168,504).  Each method is run once with seed~0.


\paragraph{Compared methods.}
We compare RIAG with eight multi-agent methods: Debate
\citep{du2023multiagentdebate}, Self-Consistency \citep{wang2023selfconsistency},
Refine \citep{madaan2023selfrefine}, ReConcile \citep{chen2024reconcile},
Mixture-of-Agents \citep{wang2024mixtureofagents}, Self-MoA \citep{li2025selfmoa},
Graph-of-Agents \citep{yun2026graphofagents}, and CIA \citep{wu2026cia}.
We run Graph-of-Agents (GoA) with max or mean pooling and homogeneous or
heterogeneous pools.  CIA originally infers communication topology; our
answer-generation adaptation uses its dependency graph for a second answering
pass, totaling 26 calls/query.  Following GoA, all methods use six
Qwen2.5-7B-Instruct agents except heterogeneous GoA, whose agents span general,
code, math, biomedical, finance, and legal models.

\paragraph{Metrics.}
We report \emph{selection accuracy} (gold document), \emph{answer accuracy}
(accepted answer for the gold task), and \emph{joint accuracy} (both on the same
query).  Our cost measure, \emph{calls per query}, divides all recorded model
calls by scheduled queries and counts a cache hit as zero calls.

\paragraph{Settings.}
All methods use temperature 0.7, top-$p$ 1.0, and a 2,048-token output cap.
Baselines use a 16,384-token context window; GoA and RIAG use 32,768 tokens.

\subsection{Main results}

\paragraph{Effectiveness.}
Table~\ref{tab:overall} compares all methods on the full collection.  Debate is
the strongest comparison method at 32.22\% joint accuracy.  Comparison methods
select the gold document on about 60\% of queries but remain below 33\% joint
accuracy; their answer accuracy is also about seven points higher than joint
accuracy, showing that some accepted answers come from the wrong document.
Homogeneous and heterogeneous RIAG reach 54.24\% and 49.78\% joint accuracy with
100.00\% and 98.81\% selection accuracy.  The homogeneous pool therefore
outperforms the four-model heterogeneous pool.

\paragraph{Efficiency.}
RIAG uses 0.62 and 0.63 calls/query, about 29 times fewer than Debate (18.00) and
42 times fewer than CIA (26.00).  The reduction comes from reusing solved
documents and expanding the graph only after disagreement or failed validation.
Table~\ref{tab:call-cost} gives the total number of model calls for every method.

\begin{table}[htbp]
\centering
\scriptsize
\setlength{\tabcolsep}{3pt}
\renewcommand{\arraystretch}{1.08}
\caption{Model calls on the full collection.  The last column divides each method's calls by those of homogeneous RIAG.}\label{tab:call-cost}
\begin{tabularx}{\linewidth}{@{}ll*{3}{>{\centering\arraybackslash}X}@{}}
\toprule
 &  & \textbf{Total calls $\downarrow$} & \textbf{Calls/query $\downarrow$} & \textbf{Relative to RIAG $\downarrow$}\\
\midrule
 & Debate & 6,721,704 & 18.00 & 28.90$\times$\\
 & Self-Consistency & 2,240,568 & 6.00 & 9.63$\times$\\
 & Refine & 7,095,132 & 19.00 & 30.50$\times$\\
 & ReConcile & 6,721,704 & 18.00 & 28.90$\times$\\
 & Mixture-of-Agents & 7,095,132 & 19.00 & 30.50$\times$\\
 & Self-MoA & 2,613,996 & 7.00 & 11.24$\times$\\
\multirow{-7}{*}{\textbf{Baselines}} & CIA (adapted) & 9,709,128 & 26.00 & 41.74$\times$\\
\midrule
 & max, homogeneous & 3,951,224 & 10.58 & 16.99$\times$\\
 & mean, homogeneous & 4,689,572 & 12.56 & 20.16$\times$\\
 & max, heterogeneous & 4,276,776 & 11.45 & 18.39$\times$\\
\multirow{-4}{*}{\shortstack[l]{\textbf{Graph-of-}\\\textbf{Agents}}} & mean, heterogeneous & 5,033,926 & 13.48 & 21.64$\times$\\
\midrule
\rowcolor{gray!15} & homogeneous & \textbf{232,614} & \textbf{0.62} & \textbf{1.00$\times$}\\
\rowcolor{gray!15}\multirow{-2}{*}{\shortstack[l]{\textbf{RIAG}\\\textit{(ours)}}} & heterogeneous & \underline{234,436} & \underline{0.63} & \underline{1.01$\times$}\\
\bottomrule
\end{tabularx}
\end{table}

%

Table~\ref{tab:overall} shows the temporal pattern behind the aggregate.
Every conventional method loses 24--29 points from baseline to historical
queries, whereas homogeneous RIAG remains within 53.89--54.50\% across baseline,
changed, propagation, preservation, and historical queries.

\subsection{Parallel independent calls and recorded cost}

Every full-collection RIAG run issues the two independent first-stage requests
(solver and checker) concurrently, while audit and repair remain sequential;
this scheduling leaves the four-call budget and the exposure rule unchanged.
Table~\ref{tab:fullset-cost} records the cost of the two production runs behind
Tables~\ref{tab:overall}--\ref{tab:answer-dataset}.  Calls/query stays below
one for both pools because a cache hit issues no call; the heterogeneous pool
spends slightly more calls and input tokens per query while scoring lower
(Table~\ref{tab:overall}).  The homogeneous figures include the 34,796
recovery calls of the repaired overlay, which replaced all 20,739 parse-failed
queries of the original run; the heterogeneous pre-fix run keeps its 4,453
parse-failed queries, which are counted as incorrect.  No serial
full-collection run exists, so these runs support no serial-versus-parallel
speed-up or accuracy-equivalence claim, and the wall-clock figures are single
runs on 16 workers, reported descriptively.

\begin{table}[htbp]
\centering
\scriptsize
\setlength{\tabcolsep}{3pt}
\renewcommand{\arraystretch}{1.08}
\caption{Recorded cost of the two full-collection RIAG production runs (31,119 episodes and 373,428 scheduled queries each; 16 workers; parallel independent calls enabled).  Homogeneous totals include the 34,796 recovery calls of the repaired overlay (17.4M input and 4.9M output tokens); its wall-clock covers the production phase only.}\label{tab:fullset-cost}
\begin{tabularx}{\linewidth}{@{}l*{2}{>{\centering\arraybackslash}X}@{}}
\toprule
\textbf{Metric} & \textbf{Homogeneous} & \textbf{Heterogeneous}\\
\midrule
Model calls $\downarrow$ & 232,614 & 234,436\\
Calls/query $\downarrow$ & 0.62 & 0.63\\
Input tokens $\downarrow$ & 98,070,477 & 106,642,017\\
Output tokens $\downarrow$ & 27,697,101 & 27,486,066\\
Output tokens/query $\downarrow$ & 74.2 & 73.6\\
Parse-failed queries, original run $\downarrow$ & 20,739 & 4,453\\
Parse-failed queries, reported row $\downarrow$ & 0 & 4,453\\
Wall-clock (s) $\downarrow$ & 8,410 & 21,160\\
\bottomrule
\end{tabularx}
\end{table}

\subsection{Interpretation and limitations}
\label{app:limits-detailed}
\label{sec:limits}

These experiments evaluate RIAG as a complete system, not the isolated effect
of its graph policy.  RIAG combines deterministic resolution, episode-local
reuse, its model configuration, and a repaired answer path; comparison runs
also differ in decoding constraints and context length.  The full collection
uses one seed, checkpoints within an episode are correlated, and MATH uses exact
string scoring.  Isolating the mechanisms requires matched resolver and cache
controls, multiple inference seeds, and randomized post-trigger actions.


\subsection{Dataset and query-category breakdowns}
\label{appedix:perf_breakdown}

Tables~\ref{tab:joint-dataset}, \ref{tab:answer-dataset},
and~\ref{tab:selection-dataset} report joint, answer, and selection accuracy by
dataset.  On MATH, homogeneous and heterogeneous RIAG
reach 63.20\% and 54.50\% joint accuracy; no conventional baseline exceeds
29.60\%.  This gap does not identify graph structure: the homogeneous row uses
answer canonicalization, and resolver, multiple-choice decoding, and context
length differ across methods.  Still, the metric gap is informative.  RIAG
selects correctly on 100.00\% and 98.81\% of queries but answers only 54.24\%
and 49.78\% correctly; non-RIAG methods select correctly on 58.68--63.51\% while
answering 33.78--39.42\%.  Better routing alone does not solve the task.

\begin{table}[htbp]
\centering
\scriptsize
\setlength{\tabcolsep}{3pt}
\renewcommand{\arraystretch}{1.08}
\caption{Joint accuracy by dataset on the full collection; each column uses every query of its dataset (GPQA 2,376; MATH 6,000; HumanEval 1,968; MedMCQA 50,196; MMLU-Pro 144,384; MMLU 168,504).}\label{tab:joint-dataset}
\begin{tabularx}{\linewidth}{@{}ll*{7}{>{\centering\arraybackslash}X}@{}}
\toprule
 &  & \multicolumn{3}{c}{Multi-Domain $\uparrow$} & \multicolumn{3}{c}{Domain-Specific $\uparrow$} & \multicolumn{1}{c}{}\\
\cmidrule(lr){3-5}\cmidrule(lr){6-8}
 &  & \textbf{MMLU} & \textbf{MMLU-Pro} & \textbf{GPQA} & \textbf{MATH} & \textbf{HumanEval} & \textbf{MedMCQA} & \textbf{All $\uparrow$}\\
\midrule
 & Debate & 38.66 & 25.76 & 23.02 & 29.60 & 14.53 & 30.66 & 32.22\\
 & Self-Consistency & 34.28 & 18.62 & 21.51 & 2.93 & 25.41 & 27.02 & 26.62\\
 & Refine & 34.58 & 20.73 & 18.73 & 20.93 & 22.26 & 27.14 & 27.84\\
 & ReConcile & 35.44 & 21.61 & 18.77 & 27.60 & 26.88 & 28.80 & 28.92\\
 & Mixture-of-Agents & 35.85 & 21.70 & 18.77 & 25.77 & 22.36 & 27.98 & 28.98\\
 & Self-MoA & 33.52 & 18.37 & 17.21 & 11.95 & 19.92 & 26.32 & 26.17\\
\multirow{-7}{*}{\textbf{Baselines}} & CIA (adapted) & 36.37 & 22.56 & 21.30 & 15.18 & 12.40 & 28.09 & 29.36\\
\midrule
 & max, homogeneous & 34.33 & 21.60 & 18.10 & 27.60 & 12.50 & 27.60 & 28.18\\
 & mean, homogeneous & 35.28 & 22.04 & 18.69 & 26.65 & 18.70 & 28.64 & 28.94\\
 & max, heterogeneous & 34.58 & 22.59 & 18.22 & 27.23 & 9.76 & 27.62 & 28.65\\
\multirow{-4}{*}{\shortstack[l]{\textbf{Graph-of-}\\\textbf{Agents}}} & mean, heterogeneous & 34.56 & 22.45 & 19.07 & 26.73 & 17.38 & 27.97 & 28.68\\
\midrule
\rowcolor{gray!15} & homogeneous & \textbf{66.83} & \textbf{40.10} & \textbf{40.40} & \textbf{63.20} & \textbf{64.02} & \textbf{51.87} & \textbf{54.24}\\
\rowcolor{gray!15}\multirow{-2}{*}{\shortstack[l]{\textbf{RIAG}\\\textit{(ours)}}} & heterogeneous & \underline{61.65} & \underline{36.89} & \underline{33.33} & \underline{54.50} & \underline{49.85} & \underline{47.25} & \underline{49.78}\\
\bottomrule
\end{tabularx}
\end{table}

\begin{table}[t]
\centering
\scriptsize
\setlength{\tabcolsep}{3pt}
\renewcommand{\arraystretch}{1.08}
\caption{Answer accuracy by dataset on the full collection; each column uses every query of its dataset (GPQA 2,376; MATH 6,000; HumanEval 1,968; MedMCQA 50,196; MMLU-Pro 144,384; MMLU 168,504).  The RIAG rows equal their joint rows because the RIAG artifacts record one exact-correctness predicate.}\label{tab:answer-dataset}
\begin{tabularx}{\linewidth}{@{}ll*{7}{>{\centering\arraybackslash}X}@{}}
\toprule
 &  & \multicolumn{3}{c}{Multi-Domain $\uparrow$} & \multicolumn{3}{c}{Domain-Specific $\uparrow$} & \multicolumn{1}{c}{}\\
\cmidrule(lr){3-5}\cmidrule(lr){6-8}
 &  & \textbf{MMLU} & \textbf{MMLU-Pro} & \textbf{GPQA} & \textbf{MATH} & \textbf{HumanEval} & \textbf{MedMCQA} & \textbf{All $\uparrow$}\\
\midrule
 & Debate & 48.41 & 29.60 & 31.99 & 30.05 & 14.53 & 39.92 & 39.42\\
 & Self-Consistency & 44.79 & 22.80 & 30.56 & 3.17 & 25.46 & 36.60 & 34.33\\
 & Refine & 44.81 & 24.70 & 26.73 & 21.33 & 22.31 & 36.70 & 35.34\\
 & ReConcile & 44.25 & 24.89 & 26.30 & 28.00 & 26.88 & 37.33 & 35.37\\
 & Mixture-of-Agents & 45.96 & 25.45 & 27.06 & 26.22 & 22.56 & 37.80 & 36.37\\
 & Self-MoA & 43.90 & 22.33 & 25.42 & 12.32 & 19.92 & 36.26 & 33.78\\
\multirow{-7}{*}{\textbf{Baselines}} & CIA (adapted) & 46.52 & 26.78 & 31.52 & 15.60 & 12.40 & 37.51 & 36.90\\
\midrule
 & max, homogeneous & 43.61 & 25.02 & 27.78 & 28.13 & 12.50 & 36.83 & 34.99\\
 & mean, homogeneous & 45.51 & 26.05 & 28.37 & 27.23 & 18.75 & 38.27 & 36.47\\
 & max, heterogeneous & 44.50 & 26.35 & 27.86 & 27.68 & 9.76 & 37.15 & 35.94\\
\multirow{-4}{*}{\shortstack[l]{\textbf{Graph-of-}\\\textbf{Agents}}} & mean, heterogeneous & 44.90 & 26.56 & 29.59 & 27.25 & 17.38 & 37.85 & 36.33\\
\midrule
\rowcolor{gray!15} & homogeneous & \textbf{66.83} & \textbf{40.10} & \textbf{40.40} & \textbf{63.20} & \textbf{64.02} & \textbf{51.87} & \textbf{54.24}\\
\rowcolor{gray!15}\multirow{-2}{*}{\shortstack[l]{\textbf{RIAG}\\\textit{(ours)}}} & heterogeneous & \underline{61.65} & \underline{36.89} & \underline{33.33} & \underline{54.50} & \underline{49.85} & \underline{47.25} & \underline{49.78}\\
\bottomrule
\end{tabularx}
\end{table}

\begin{table}[t]
\centering
\scriptsize
\setlength{\tabcolsep}{3pt}
\renewcommand{\arraystretch}{1.08}
\caption{Selection accuracy by dataset on the full collection.}\label{tab:selection-dataset}
\begin{tabularx}{\linewidth}{@{}ll*{7}{>{\centering\arraybackslash}X}@{}}
\toprule
 &  & \multicolumn{3}{c}{Multi-Domain $\uparrow$} & \multicolumn{3}{c}{Domain-Specific $\uparrow$} & \multicolumn{1}{c}{}\\
\cmidrule(lr){3-5}\cmidrule(lr){6-8}
 &  & \textbf{MMLU} & \textbf{MMLU-Pro} & \textbf{GPQA} & \textbf{MATH} & \textbf{HumanEval} & \textbf{MedMCQA} & \textbf{All $\uparrow$}\\
\midrule
 & Debate & 62.08 & 63.28 & 62.75 & 63.17 & 51.27 & 64.13 & 62.78\\
 & Self-Consistency & 61.07 & 61.67 & 59.43 & 58.62 & 54.27 & 62.47 & 61.40\\
 & Refine & 60.70 & 60.88 & 59.26 & 59.32 & 45.53 & 62.09 & 60.85\\
 & ReConcile & 62.57 & 64.32 & 62.79 & \underline{64.30} & 64.43 & 64.29 & 63.51\\
 & Mixture-of-Agents & 60.85 & 61.26 & 59.93 & 59.80 & 42.73 & 62.11 & 61.06\\
 & Self-MoA & 60.85 & 61.22 & 59.64 & 58.07 & 36.99 & 61.83 & 60.95\\
\multirow{-7}{*}{\textbf{Baselines}} & CIA (adapted) & 61.27 & 62.05 & 60.61 & 59.27 & 46.19 & 63.09 & 61.70\\
\midrule
 & max, homogeneous & 60.03 & 61.23 & 57.62 & 61.12 & 52.03 & 60.99 & 60.58\\
 & mean, homogeneous & 58.92 & 58.84 & 57.45 & 59.32 & 56.35 & 61.14 & 59.17\\
 & max, heterogeneous & 59.52 & 60.71 & 57.07 & 60.68 & 51.47 & 61.64 & 60.23\\
\multirow{-4}{*}{\shortstack[l]{\textbf{Graph-of-}\\\textbf{Agents}}} & mean, heterogeneous & 58.17 & 58.62 & 56.73 & 58.30 & 51.37 & 61.03 & 58.68\\
\midrule
\rowcolor{gray!15} & homogeneous & \textbf{100.00} & \textbf{100.00} & \textbf{100.00} & \textbf{100.00} & \textbf{100.00} & \textbf{100.00} & \textbf{100.00}\\
\rowcolor{gray!15}\multirow{-2}{*}{\shortstack[l]{\textbf{RIAG}\\\textit{(ours)}}} & heterogeneous & \underline{99.39} & \underline{97.97} & \underline{94.36} & \textbf{100.00} & \underline{85.57} & \underline{99.84} & \underline{98.81}\\
\bottomrule
\end{tabularx}
\end{table}

\subsection{MATH Analysis}
\label{sec:full-math-analysis}

We now examine the MATH sweep, which is separate from the six-dataset comparison
in Section~\ref{sec:experiments}.  Nine one-factor settings are crossed with
homogeneous and heterogeneous pools, giving 18 configurations, 500 episodes per
configuration, and 108,000 scheduled queries.  The reference uses temperature
0.7, top-$p$ 1.0, a 2,048-token output cap, and seed~0; each alternative changes
one setting.  Seed changes measure inference sensitivity, not independent
training replications.

These executions also differ from the MATH runs in
Table~\ref{tab:joint-dataset}: that table uses the repaired homogeneous answer
path and the original pre-fix heterogeneous run.  Accordingly, reference
accuracy here is 64.13\% rather than 63.20\% for the homogeneous pool and 54.88\%
rather than 54.50\% for the heterogeneous pool.  We do not pool the two studies.

Accuracy differences are paired by episode against the corresponding reference.
Percentile 95\% intervals use 10,000 bootstrap resamples of 500 episodes, keeping
each episode's 12 queries together.  All runs finish without query or model-call
errors.  We retain the runner's exact-string joint metric rather than substitute
symbolic equivalence.

\subsubsection{Resolver-controlled Graph-of-Agents}

Figure~\ref{fig:goa-math-resolver-control} in the main text tests whether the
resolver-controlled interface extends beyond RIAG.  Eight MATH configurations
cross max or mean pooling, homogeneous or heterogeneous pools, and vanilla or
resolver-controlled GoA.  Each covers the same 500 episodes and 6,000 queries.
The controlled variant applies the public deterministic resolver before GoA's
reasoning graph and does not cache answers.  Its answer contract supports
constrained option decoding, but MATH is free-form; the constraint is inactive
and this study provides no evidence about its effect.

The transferred design produces the largest change in task binding: selection
rises from 58.30--61.12\% for vanilla GoA to 99.35--99.87\% with the resolver.
Joint accuracy rises from 27.60\% to 57.67\% for homogeneous
max pooling and from 26.65\% to 57.28\% for homogeneous mean pooling, gains of
30.07 and 30.63 points.  Heterogeneous max gains 19.95 points (27.23\% to
47.18\%); heterogeneous mean gains 27.73 points (26.73\% to 54.47\%).  Similar
answer-accuracy gains (19.50--30.05 points) show why routing and solution quality
must be reported separately.

Cost does not increase uniformly: resolver-controlled homogeneous runs use
0.60--0.68 more calls/query, while heterogeneous runs use 0.17--0.25 fewer.
Parse failures fall from 369 to 86 across the four runs, but the inactive
multiple-choice constraint cannot explain that reduction.  Pooling matters most
for controlled heterogeneous GoA, where mean exceeds max by 7.28 joint-accuracy
points; the homogeneous variants differ by only 0.38 points.

The consistent direction across all four GoA settings supports architectural
generality: temporal resolution can be composed with a reasoning graph other
than RIAG's.  This is not, however, a one-factor ablation.  Resolver-controlled runs also use four fixed
role cards and greedy decoding, whereas vanilla GoA uses its six-agent reference
pool at temperature 0.7.  With one run per cell and no episode-level outcomes
for paired uncertainty, the observed gap belongs to the complete
resolver-controlled package.  Isolating the resolver requires a fully crossed
experiment over answer contract, agent pool, and decoding temperature.

\subsubsection{Sensitivity to decoding and output budget}

\begin{figure}[t]
\centering
\includegraphics[width=\linewidth]{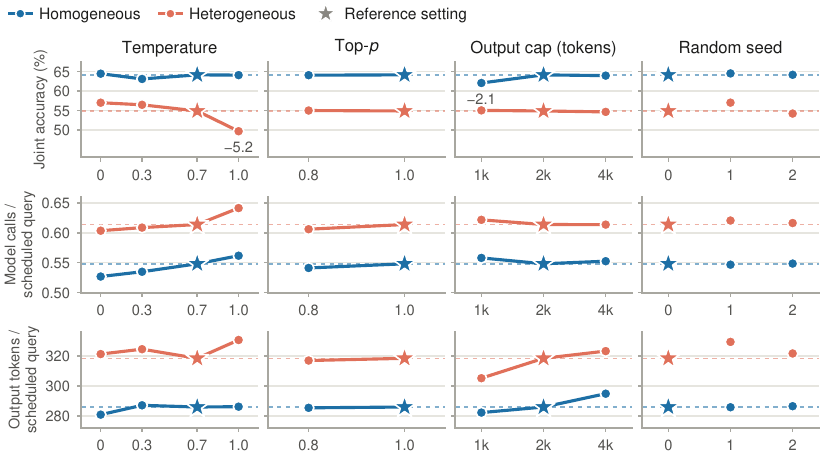}
\caption{RIAG on the MATH split with one setting varied at a time (500 episodes, 6,000 queries per run); stars mark the reference.  Rows: joint accuracy, model calls per query, and output tokens per query.  Labels give the two accuracy changes whose paired 95\% intervals exclude zero.}
\label{fig:math-efficiency}
\end{figure}
Figure~\ref{fig:math-efficiency} next tests local robustness.  Homogeneous RIAG
ranges from 62.05\% to 64.50\% across the sweep; heterogeneous RIAG ranges from
49.70\% to 57.02\%.  Most paired intervals include zero.  Two do not: a
1,024-token homogeneous cap changes accuracy by $-2.08$ points (95\% CI
$[-3.85,-0.33]$), and heterogeneous temperature 1.0 changes it by $-5.18$
points ($[-7.83,-2.58]$).  The shorter homogeneous cap produces 115 format
failures, versus 39 for the heterogeneous 1,024-token condition and four for
heterogeneous seed~1; all other cells have none.  The temperature effect occurs
without format failures, indicating sampling sensitivity rather than truncation.

Greedy decoding has positive point estimates in both pools ($+0.30$ homogeneous,
$+2.12$ heterogeneous), but both intervals include zero; so does the apparent
heterogeneous gain at seed~1.  Thus, the homogeneous pool is stable over most
tested settings, but neither pool is invariant to output budget or sampling.
The one-factor design does not identify interactions.


\subsubsection{Accuracy and recorded compute}

The reference homogeneous pool reaches 64.13\% accuracy at 0.548 calls and
285.9 output tokens/query; the heterogeneous pool reaches 54.88\% at 0.614 calls
and 318.5 tokens/query.  Here, heterogeneity costs more and performs worse.

The lower rows of Figure~\ref{fig:math-efficiency} relate cost to accuracy.
Greedy homogeneous decoding uses the fewest calls and is slightly more accurate
than its reference, although the interval includes zero.  A 1,024-token
homogeneous cap reduces tokens and accuracy but increases calls/q.  For the
heterogeneous pool, temperature 1.0 produces the lowest accuracy and the highest
call and token counts; a larger output cap adds tokens without a matching gain.
Calls/q also rise with temperature in both pools, consistent with more initial
disagreement and hence more audits, though we do not test that mechanism
directly.  These observations guide operating-point choice, not a scaling law.

\subsubsection{Temporal reuse and conditional graph expansion}

\begin{figure}[t]
  \centering
  \includegraphics[width=\linewidth]{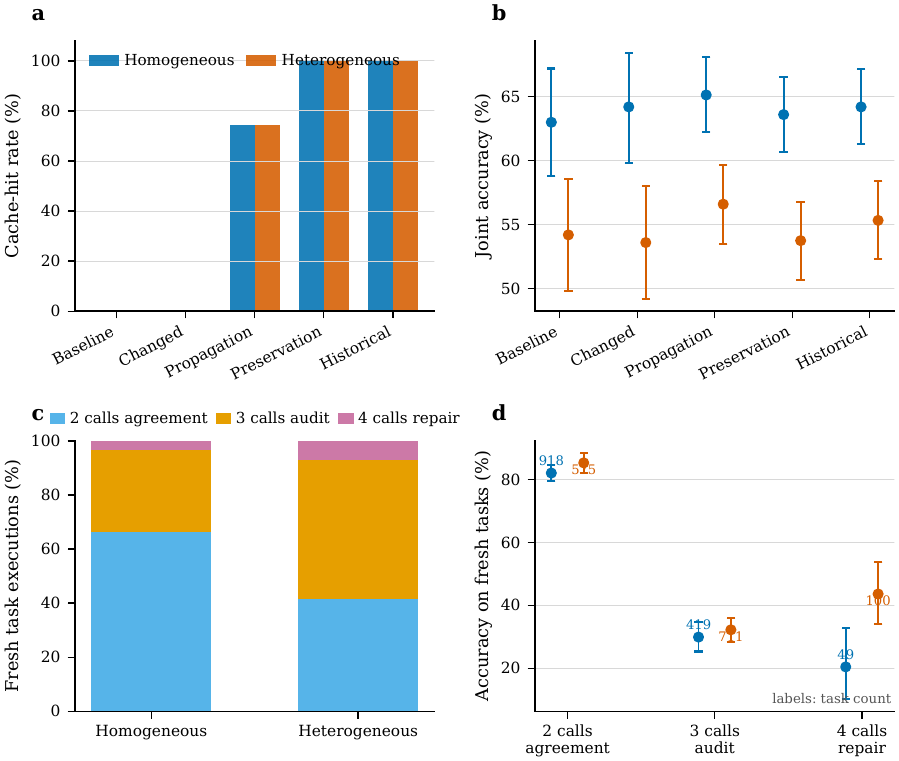}
  \caption{Reference-setting mechanism diagnostics.  (a) Cache-hit rate by
  query category.  (b) Joint accuracy by category with episode-cluster
  bootstrap intervals.  (c) Route depth among unique fresh task executions.
  (d) Route-conditioned accuracy; numeric labels give task counts.  Panels
  (c,d) describe the policy's selection behavior and are not causal effects of
  additional calls.}
  \label{fig:math-mechanisms}
\end{figure}

Temporal reuse explains most of the low calls/q.  In both reference pools,
4,614 of 6,000 queries (76.9\%) are cache hits.  Baseline and changed queries
always trigger fresh reasoning; 74.3\% of propagation queries and all
preservation and historical queries reuse a solution.  In
Figure~\ref{fig:math-mechanisms}, accuracy spans 63.0--65.1\% across categories
for homogeneous RIAG and 53.6--56.6\% for heterogeneous RIAG.  This stability is
consistent with successful reuse, but its causal value requires a matched
no-cache control.

The 1,386 fresh tasks expose conditional graph expansion.  Homogeneous RIAG
stops after two agreeing calls on 66.2\%, invokes the auditor on 30.2\%, and
reaches repair on 3.5\%.  The heterogeneous rates are 41.5\%, 51.3\%, and 7.2\%,
showing more disagreement and deeper graphs.  Agreement-route accuracy is high
in both pools (82.1\% and 85.4\%), while audited tasks are harder.  These
conditional accuracies are not intervention effects: disagreement or failed
validation selects tasks into deeper routes.  A causal estimate must compare
alternate actions on the same triggered tasks.

\subsubsection{Reasoning-space diversity and conditional intervention}
\label{sec:math-reasoning-umap}

Graph depth does not show whether independent derivations differ substantively.
We therefore analyze 1{,}386 matched fresh MATH tasks per reference
configuration: 7{,}004 invocations (3{,}298 homogeneous and 3{,}706
heterogeneous), excluding cache hits.  We embed valid JSON reason fields with
the MiniLM encoder from Appendix~\ref{sec:dataset-umap}; malformed responses use
their full text.  Numerical comparisons use the original normalized
384-dimensional embeddings, while UMAP serves only as a shared display.

For display, we remove task-subject variation by subtracting each task's mean
embedding across roles and configurations, then renormalizing.  The joint UMAP
uses cosine distance, 50 neighbors, \texttt{min\_dist}${=}0.15$, and seed~42.
Its 15-neighbor trustworthiness is 0.728, versus 0.946 for the raw-vector map;
we therefore avoid claims based on projected local detail.

Figure~\ref{fig:riag-role-umap} shows detectable role and stage signatures even
with a fixed checkpoint.  A four-class role probe with five-fold,
episode-grouped cross-validation reaches 34.3\% balanced accuracy for
homogeneous RIAG (25\% chance) and 65.7\% for heterogeneous RIAG, where role and
checkpoint are confounded.  Initial solver/checker responses predict the pool
at 67.5\% balanced accuracy, so model assignment changes response geometry
before communication.

\begin{figure}[t]
  \centering
  \includegraphics[width=0.95\linewidth]{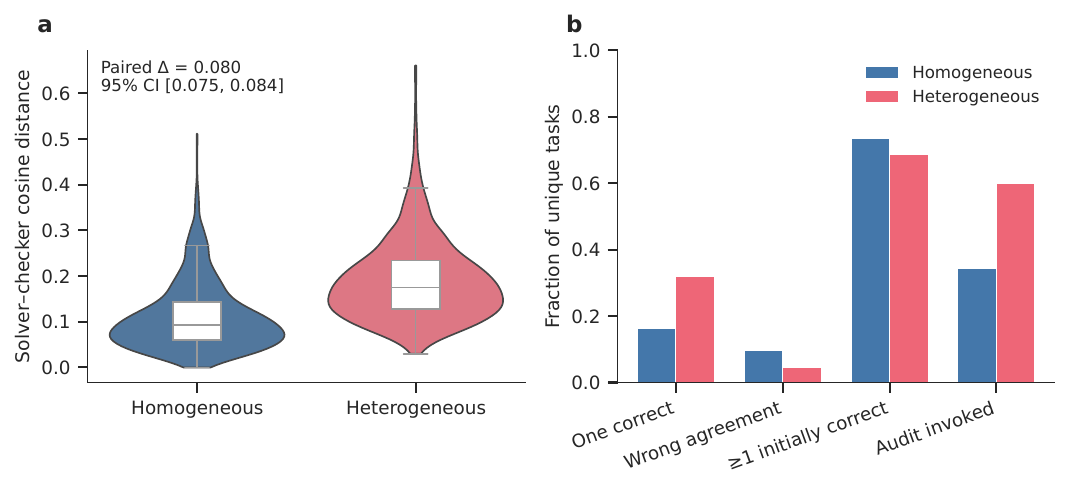}
  \caption{Matched solver--checker diversity on 1{,}386 unique MATH tasks per
  configuration.  (a) Original-space cosine-distance distributions; the
  annotation reports the paired heterogeneous-minus-homogeneous difference and
  a 10{,}000-resample task-bootstrap interval.  (b) Initial-pair outcomes and
  audit frequency.  ``One correct'' denotes useful diversity; ``wrong
  agreement'' denotes equal normalized answers that are both incorrect.}
  \label{fig:riag-pair-diversity}
\end{figure}

Figure~\ref{fig:riag-pair-diversity} compares solver--checker diversity in the
original embedding space.  Mean cosine distance is 0.190 for heterogeneous and
0.111 for homogeneous RIAG.  The paired difference is 0.080 (95\% CI
$[0.075,0.084]$), with larger heterogeneous distances on 83.8\% of tasks and a
standardized effect of 0.85.  This establishes response diversity, not better
reasoning.

The error structure explains the qualification.  Heterogeneity raises useful
diversity from 16.4\% to 32.0\% and lowers wrong agreement from 9.5\% to 4.6\%,
yet initial answer coverage falls from 73.4\% to 68.8\%.  Final normalized
accuracy on unique tasks is therefore 67.6\% for homogeneous and 57.5\% for
heterogeneous RIAG.  This diagnostic weights each fresh task once and uses
normalized strings; it does not replace scheduled-query exact match.

\begin{figure}[t]
  \centering
  \includegraphics[width=\linewidth]{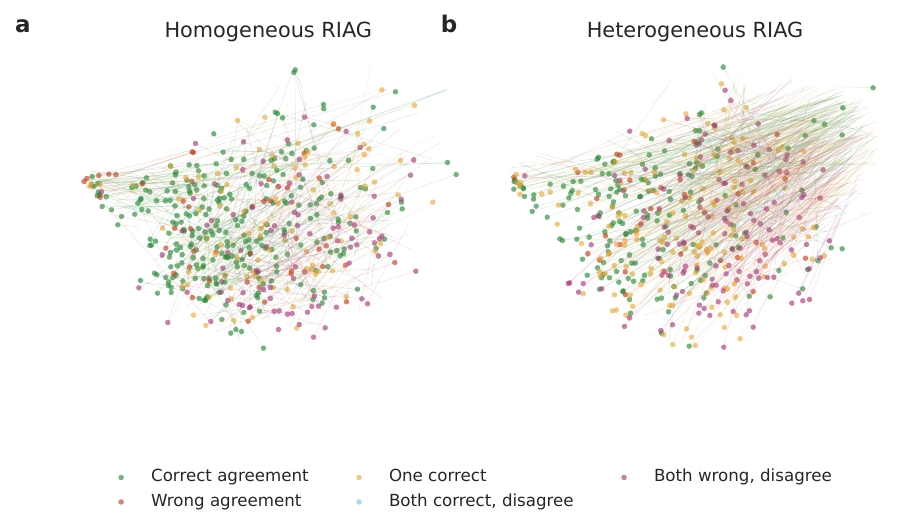}
  \caption{Initial-pair outcomes on a fixed display subset of the task-centered
  UMAP.  Each segment links the solver and checker responses for one task, and
  the solver endpoint is colored by normalized answer outcome.  All matched
  pairs enter the numerical analysis; the plotted subset is for readability
  and is not a learned decision boundary.}
  \label{fig:riag-pair-outcomes}
\end{figure}

Figure~\ref{fig:riag-pair-outcomes} visualizes pair outcomes without using 2-D
segment length or apparent clusters for inference.  In the original space,
wrong agreements are closer than disagreeing pairs in both pools, and
heterogeneous ``both wrong, disagree'' pairs are the most separated group.
Distance therefore signals disagreement but not whether it is productive.

Useful diversity rises with pair distance in
Figure~\ref{fig:riag-distance-audit}, but coverage falls.  From the closest to
farthest quartile, initial answer coverage drops from 80.7\% to 62.8\% for
homogeneous RIAG and from 79.8\% to 55.9\% for heterogeneous RIAG.  Distance
alone rewards novelty without ensuring a correct candidate.

Auditing is invoked on 478 homogeneous and 830 heterogeneous tasks, including
graphs that later reach repair.  Conditional rescue rates are 2.5\% and 0.6\%,
damage rates 19.2\% and 19.4\%, and final normalized accuracies 32.8\% and
36.5\%.  Because disagreement or failed validation triggers auditing, these are
descriptive rates, not causal effects; causal estimation requires randomizing
the post-trigger action.

Overall, heterogeneous role--model assignment produces more distinct and more
complementary answers, but diversity neither guarantees a correct candidate nor
improves final accuracy.  Separating role from model requires a full factorial
design.  The embedding analysis also captures style, truncates late derivation
steps, and inherits normalized-string scoring errors for equivalent expressions.

\paragraph{Scope of the evidence.}
The sweep supports bounded execution, temporal reuse, and robustness near the
reference setting.  It does not isolate the resolver, cache, graph policy, and
model mixture; provide matched cross-method controls; or test hyperparameter
interactions.  Exact-string scoring can also reject equivalent expressions.
The figures therefore provide mechanism and sensitivity evidence, not a causal
decomposition of RIAG's aggregate advantage.

\section{Dynamic Task Routing Collection: Curation Protocol and Statistics}
\label{app:dataset}

\subsection{Curation and review protocol}
\label{sec:dataset-curation}

The collection begins from local snapshots of these six public benchmarks. We freeze the
byte-level source files, their order, local split names, counts, and SHA-256
hashes before curation. Astra was used as an AI curation
assistant while developing the event-stream transformation and inspecting
candidate presentations.  It did not replace source questions, options, answer
keys, HumanEval entry points, or hidden tests, and it was not used as a source
of gold labels.  The released 31{,}119 episodes are authoritatively
materialized by the seeded Python builder described in
\path{datasets/dynamic_event_stream/REPRODUCIBLE_PROTOCOL.md}; thus rebuilding
from the frozen exports, recorded Python version, seed, and script snapshot
does not require a model call.

After construction, three human reviewers examined samples from the resulting
datasets.  Review covered (i) fidelity of each public task document to its
mapped source row, (ii) preservation of answer options and code interfaces,
(iii) coherence of arrival and effective times, supersession, expiry, and
retraction, (iv) agreement between the selected document and private scoring
record, and (v) absence of gold answers, tests, future events, and private
metadata from the public stream.

Table~\ref{tab:curation-protocol} summarizes the construction stages and the
control applied at each stage.

\begin{table}[t]
\centering
\scriptsize
\setlength{\tabcolsep}{3pt}
\renewcommand{\arraystretch}{1.10}
\caption{Dataset curation and quality-control stages.  AI assistance supports
protocol development, while deterministic generation and independent replay
define the released artifacts and their gold bindings.}
\label{tab:curation-protocol}
\begin{tabular}{@{}p{0.93in}p{2.05in}>{\raggedright\arraybackslash}p{2.34in}@{}}
\toprule
\textbf{Stage} & \textbf{Operation} & \textbf{Control} \\
\midrule
Source freeze & Snapshot the six public benchmarks also evaluated by
Graph-of-Agents. & Preserve source bytes, row order, split labels, counts, and
hashes. \\
AI-assisted design & Use Astra to help develop and inspect the dynamic
event-stream transformation. & Do not use model output to rewrite source tasks
or establish gold answers. \\
Deterministic build & Map every source row to one episode; sample two
within-dataset partners; generate 12 temporal queries. & Seeded standard-library
Python, frozen script snapshot, immutable source content, and private gold. \\
Human review & Three reviewers inspect samples from all resulting dataset
families. & Check source fidelity, temporal semantics, document--gold binding,
and public/private separation. \\
Mechanical audit & Replay temporal selection with an independent validator and
run regression tests. & Verify 31{,}119 episodes, 373{,}428 queries, schemas,
manifests, and artifact checksums. \\
\bottomrule
\end{tabular}
\end{table}

Table~\ref{tab:episode-examples} expands a generated episode from each
dataset rather than compressing an episode into a label sequence.  Each block
first maps document labels to exact source rows and abbreviated public tasks,
then reports every checkpoint.  Here $a$ is the time at which the query is
issued and $w$ is the world time about which it asks; an effective time $e$ is
shown in the update column when it differs from arrival time.  ``Selected'' is
the private gold document binding, not an answer key or HumanEval test.

{\scriptsize
\setlength{\tabcolsep}{2pt}
\renewcommand{\arraystretch}{1.08}
\begin{longtable}{@{}r>{\raggedright\arraybackslash}p{2.85in}rr>{\raggedright\arraybackslash}p{0.88in}c@{}}
\caption{Six actual episodes.}
\label{tab:episode-examples}\\
\toprule
\textbf{\#} & \textbf{Public update before the query} & \textbf{$a$} &
\textbf{$w$} & \textbf{Checkpoint} & \textbf{Selected} \\
\midrule
\endfirsthead
\multicolumn{6}{l}{\tablename~\thetable\ (continued)}\\
\toprule
\textbf{\#} & \textbf{Public update before the query} & \textbf{$a$} &
\textbf{$w$} & \textbf{Checkpoint} & \textbf{Selected} \\
\midrule
\endhead
\midrule
\multicolumn{6}{r}{Continued on next page}\\
\endfoot
\bottomrule
\endlastfoot

\multicolumn{6}{@{}p{5.25in}@{}}{\textbf{MMLU --- \texttt{mmlu-task-version-000000}.}
Documents: A = row 3370 (commander analogy); B = row 11280 (new-van
liability); C = row 0 (field-extension degree).}\\
1 & Load B, A, and C; bind requests; activate C & 2 & 2 & Baseline & C\\
2 & Supersede the active binding with B & 7 & 7 & Changed & B\\
3 & Add an unrelated request; query the archive & 12 & 12 & Preservation & C\\
4 & No new update; query the earlier world state & 12 & 2 & Historical & C\\
5 & Schedule C for $e=28$ & 18 & 18 & Preservation & B\\
6 & No new update; scheduled C is now effective & 28 & 28 & Propagation & C\\
7 & Add temporary binding B & 36 & 36 & Propagation & B\\
8 & Add delayed B with $e=27$ & 43 & 43 & Preservation & B\\
9 & Expire the temporary binding & 52 & 36 & Historical & B\\
10 & No new update; query the current state & 52 & 52 & Propagation & C\\
11 & Retract the temporary binding & 57 & 36 & Historical & C\\
12 & No new update; query the archive & 66 & 66 & Preservation & C\\
\addlinespace

\multicolumn{6}{@{}p{5.25in}@{}}{\textbf{MMLU-Pro --- \texttt{mmlu-pro-task-version-000000}.}
Documents: A = row 0 (advertising regulation); B = row 9757 (relativistic
flashes); C = row 7332 (field-license dispute).}\\
1 & Load A, B, and C; bind requests; activate A & 17 & 17 & Baseline & A\\
2 & Supersede the active binding with C & 23 & 23 & Changed & C\\
3 & Schedule B for $e=36$ & 29 & 29 & Preservation & C\\
4 & No new update; scheduled B is now effective & 37 & 37 & Propagation & B\\
5 & Add temporary binding C & 39 & 39 & Propagation & C\\
6 & Add delayed C with $e=35$ & 48 & 48 & Preservation & C\\
7 & Expire the temporary binding & 53 & 53 & Propagation & B\\
8 & No new update; query the earlier world state & 53 & 39 & Historical & C\\
9 & Retract the temporary binding & 55 & 39 & Historical & B\\
10 & Add an unrelated request; query the archive & 62 & 62 & Preservation & A\\
11 & No new update; query the initial world state & 62 & 17 & Historical & A\\
12 & No new update; query the archive & 70 & 70 & Preservation & A\\
\addlinespace

\multicolumn{6}{@{}p{5.25in}@{}}{\textbf{GPQA --- \texttt{gpqa-task-version-000000}.}
Documents: A = row 20 (structural isomers); B = row 160 (electron
microscopy); C = row 0 (quantum-state lifetimes).}\\
1 & Load B, C, and A; bind requests; activate C & 6 & 6 & Baseline & C\\
2 & Supersede the active binding with B & 14 & 14 & Changed & B\\
3 & Schedule A for $e=26$ & 18 & 18 & Preservation & B\\
4 & No new update; scheduled A is now effective & 29 & 29 & Propagation & A\\
5 & Add an unrelated request; query the initial state & 33 & 6 & Historical & C\\
6 & No new update; query the archive & 33 & 33 & Preservation & C\\
7 & Add temporary binding C & 38 & 38 & Propagation & C\\
8 & Add delayed C with $e=25$ & 44 & 44 & Preservation & C\\
9 & Expire the temporary binding & 51 & 51 & Propagation & A\\
10 & No new update; query the earlier world state & 51 & 38 & Historical & C\\
11 & Retract the temporary binding & 54 & 38 & Historical & A\\
12 & No new update; query the archive & 63 & 63 & Preservation & C\\
\addlinespace

\multicolumn{6}{@{}p{5.25in}@{}}{\textbf{MATH --- \texttt{math-task-version-000000}.}
Documents: A = row 89 (greatest common divisor); B = row 0 (polar
coordinates); C = row 405 (office assignments).}\\
1 & Load C, A, and B; bind requests; activate B & 14 & 14 & Baseline & B\\
2 & Supersede the active binding with A & 18 & 18 & Changed & A\\
3 & Add an unrelated request; query the initial state & 24 & 14 & Historical & B\\
4 & No new update; query the archive & 24 & 24 & Preservation & B\\
5 & Schedule C for $e=39$ & 31 & 31 & Preservation & A\\
6 & No new update; scheduled C is now effective & 39 & 39 & Propagation & C\\
7 & Add temporary binding A & 48 & 48 & Propagation & A\\
8 & Add delayed B with $e=38$ & 53 & 53 & Preservation & A\\
9 & Expire the temporary binding & 59 & 59 & Propagation & C\\
10 & No new update; query the earlier world state & 59 & 48 & Historical & A\\
11 & Retract the temporary binding & 64 & 48 & Historical & C\\
12 & No new update; query the archive & 71 & 71 & Preservation & B\\
\addlinespace

\multicolumn{6}{@{}p{5.25in}@{}}{\textbf{HumanEval --- \texttt{humaneval-task-version-000000}.}
Documents: A = row 25 (\texttt{factorize}); B = row 93 (\texttt{encode});
C = row 0 (\texttt{has\_close\_elements}).}\\
1 & Load A, C, and B; bind requests; activate C & 11 & 11 & Baseline & C\\
2 & Supersede the active binding with A & 19 & 19 & Changed & A\\
3 & Add an unrelated request; query the archive & 25 & 25 & Preservation & C\\
4 & No new update; query the initial world state & 25 & 11 & Historical & C\\
5 & Schedule C for $e=37$ & 34 & 34 & Preservation & A\\
6 & No new update; scheduled C is now effective & 37 & 37 & Propagation & C\\
7 & Add temporary binding B & 39 & 39 & Propagation & B\\
8 & Add delayed B with $e=36$ & 43 & 43 & Preservation & B\\
9 & Expire the temporary binding & 52 & 39 & Historical & B\\
10 & No new update; query the current state & 52 & 52 & Propagation & C\\
11 & Retract the temporary binding & 59 & 39 & Historical & C\\
12 & No new update; query the archive & 63 & 63 & Preservation & C\\
\addlinespace

\multicolumn{6}{@{}p{5.25in}@{}}{\textbf{MedMCQA --- \texttt{medmcqa-task-version-000000}.}
Documents: A = row 1932 (intertrochanteric fracture); B = row 0
(myelinated fibers); C = row 3073 (mixed-dentition analysis).}\\
1 & Load B, A, and C; bind requests; activate B & 5 & 5 & Baseline & B\\
2 & Supersede the active binding with A & 9 & 9 & Changed & A\\
3 & Add temporary binding C & 13 & 13 & Propagation & C\\
4 & Add delayed A with $e=8$ & 20 & 20 & Preservation & C\\
5 & Expire the temporary binding & 25 & 25 & Propagation & A\\
6 & No new update; query the earlier world state & 25 & 13 & Historical & C\\
7 & Retract the temporary binding & 31 & 13 & Historical & A\\
8 & Add an unrelated request; query the archive & 40 & 40 & Preservation & B\\
9 & No new update; query the initial world state & 40 & 5 & Historical & B\\
10 & Schedule C for $e=50$ & 42 & 42 & Preservation & A\\
11 & No new update; scheduled C is now effective & 53 & 53 & Propagation & C\\
12 & No new update; query the archive & 58 & 58 & Preservation & B\\
\end{longtable}
}

\subsection{Source coverage and episode construction}

We repurpose six public benchmarks also evaluated by Graph-of-Agents
\citep{yun2026graphofagents}: MMLU \citep{hendrycks2021mmlu}, MMLU-Pro
\citep{wang2024mmlupro}, MedMCQA \citep{pal2022medmcqa}, MATH
\citep{hendrycks2021math}, GPQA \citep{rein2024gpqa}, and HumanEval
\citep{chen2021humaneval} into dynamic task routing datasets.  Using one
published conversion protocol across all six benchmarks preserves their
original task content while providing consistent event semantics and
comparable routing tests across domains and answer formats.

The full schema-2.0, seed-0 collection contains every row of six named local
exports: MMLU, MMLU-Pro, MedMCQA, MATH, GPQA, and HumanEval.  This scope is the
local evaluation export, not every upstream split.  As shown in
Table~\ref{tab:routing-datasets}, the collection has 31{,}119 source items and
the same number of episodes.  MMLU and MMLU-Pro account for 83.8\% of episodes,
so aggregate scores should always be accompanied by per-dataset results.

\begin{table}[H]
\centering
\scriptsize
\setlength{\tabcolsep}{3pt}
\renewcommand{\arraystretch}{1.08}
\caption{The dynamic task routing collection.  Every source item anchors one
episode and every episode contains 12 checkpoint queries.}
\label{tab:routing-datasets}
\begin{tabularx}{\linewidth}{@{}l*{4}{>{\centering\arraybackslash}X}@{}}
\toprule
\textbf{Dataset} & \textbf{Local split} & \textbf{Source items / episodes} & \textbf{Queries} & \textbf{Episode share} \\
\midrule
MMLU & test & 14{,}042 & 168{,}504 & 45.1\% \\
MMLU-Pro & test & 12{,}032 & 144{,}384 & 38.7\% \\
MedMCQA & test & 4{,}183 & 50{,}196 & 13.4\% \\
MATH & test & 500 & 6{,}000 & 1.6\% \\
GPQA & test & 198 & 2{,}376 & 0.6\% \\
HumanEval & dev & 164 & 1{,}968 & 0.5\% \\
\midrule
Total & -- & 31{,}119 & 373{,}428 & 100.0\% \\
\bottomrule
\end{tabularx}
\end{table}

\subsection{Model inputs, output contracts, and scoring}
\label{app:io-contracts}

This subsection records the exact interfaces needed to replay the collection.
The six inputs are frozen local JSON-array exports, rather than datasets fetched
by name at evaluation time.  Table~\ref{tab:dataset-io-contracts} gives their
paths relative to the source-export root, the field copied verbatim into each
public task document, and the required answer type.  Array position is the
source row ID and the SHA-256 of the complete export is its revision; the exact
hashes and record counts are in \path{source_inventory.json}.  Re-downloading
or reserializing an upstream dataset therefore creates a different revision.

\begin{table}[H]
\centering
\scriptsize
\setlength{\tabcolsep}{3pt}
\renewcommand{\arraystretch}{1.10}
\caption{Dataset-specific input and output contracts.  The public document is
the indicated source field without rewriting.  All final predictions also
carry a document label in \texttt{document\_id}.}
\label{tab:dataset-io-contracts}
\begin{tabular}{@{}p{0.68in}p{1.48in}p{0.72in}p{0.57in}p{1.63in}@{}}
\toprule
\textbf{Dataset} & \textbf{Frozen export} & \textbf{Task field} &
\textbf{Valid answer} & \textbf{Evaluator} \\
\midrule
MMLU & \path{test/MMLU_test.json} & \texttt{question} & A--D & Exact
source option letter \\
MMLU-Pro & \path{test/MMLU_Pro_test.json} & \texttt{question} & A--J & Exact
source option letter \\
MedMCQA & \path{test/MedMCQA_test.json} & \texttt{question} & A--D & Exact
source option letter \\
GPQA & \path{test/GPQA_test.json} & \texttt{question} & A--D & Exact
source option letter \\
MATH & \path{test/MATH_test.json} & \texttt{question} & String & Exact
match to \texttt{gold\_answer} \\
HumanEval & \path{dev/human_eval_dev.json} & \texttt{prompt} & Python & Hidden
\texttt{test} with original \texttt{entry\_point} \\
\bottomrule
\end{tabular}
\end{table}

\paragraph{Public episode input.}
At \texttt{episode\_start}, every method receives the same instruction block.
It states that documents are immutable tasks rather than answer reports; that
\texttt{delivery\_request} reads \texttt{active\_task} while
\texttt{archive\_request} keeps its initial fixed binding; and that selection
uses the greatest valid
$(\textit{effective time},\textit{arrival time},\textit{event ID})$ among only
the events received so far.  It also specifies the effects of supersession,
expiry, retraction, delayed reports, future-effective assertions, and
historical queries, and explicitly says that no gold feedback is provided.
The complete block is stored once per episode in the public
\texttt{instructions} field of \path{stream.jsonl}; it is also reproduced
byte-for-byte as \texttt{INSTRUCTIONS} in the frozen
\path{source_snapshot/adapt_all_benchmarks_to_des.py}.
Here immutability is an experimental control, not a claim about all document
workflows: source-task content and its answer key remain fixed while events
change the active binding.  A content edit would require a new version, but the
current collection does not evaluate free-form document editing.
For completeness, its full text is:
\begin{quote}
\footnotesize\raggedright
This is a synthetic task-versioning workflow over real, immutable source tasks.
Task documents are not answer reports. Solve the selected document; never infer
correctness from selection. The \texttt{active\_task} property selects a task
document. The delivery request depends on \texttt{active\_task}. The archive
request permanently selects the document specified by its binding event. Return
a JSON object with \texttt{document\_id} (A, B, or C) and \texttt{answer}. If a
method wraps its response in an answer field, put this object inside that field.
The answer value is: for multiple choice, its source option letter; for
mathematics, its mathematical answer; for programming, complete Python source
implementing the selected document's original entry point. All
\texttt{active\_task} assertions have the same authority. At each query use only
events already received. At the queried world time, select the still-valid
assertion with greatest effective time (then arrival time, then ID).
Supersession closes only its target at the new effective time. Expiry closes
only its target, preserving earlier history. Retraction discredits its target
at all historical times once the retraction has arrived and become effective.
An explicitly superseded assertion never revives. Other still-valid assertions
can resume after a temporary assertion expires or is retracted. Delayed reports
do not overwrite later effective assertions. Future-effective assertions do
not apply early. Historical queries use the evidence known now, not evidence
that will arrive later. No gold feedback is given.
\end{quote}

Each source task is then delivered in the following exact form, where $L$ is
A, B, or C and the final line is the unmodified field in
Table~\ref{tab:dataset-io-contracts}:
\begin{quote}
\small\ttfamily\raggedright
Immutable task document $L$:\\
$\langle$source \texttt{question} or HumanEval \texttt{prompt}$\rangle$
\end{quote}
Multiple-choice source strings already contain their labeled options.  MATH
strings contain the original free-form problem.  HumanEval strings contain the
original function signature and docstring; solutions, tests, and private source
metadata are never placed in the public stream.

\paragraph{Public query prompts.}
Every query uses one of three templates below, followed verbatim by the common
output sentence.  Here $t$ denotes an explicitly requested historical world
time.
\begin{quote}
\small\ttfamily\raggedright
Fulfill the delivery request, which depends on active\_task at the current
world time.\\
Fulfill the delivery request, which depends on active\_task at world time $t$.\\
Fulfill the archive request at the current world time.\\[2pt]
\normalfont\textit{Common suffix:}\ttfamily\ Return a JSON object with
document\_id (A, B, or C) and answer (the source option letter, mathematical
answer string, or complete Python source).
\end{quote}
The stream exposes only \texttt{id}, \texttt{type}, \texttt{arrival\_time}, and
\texttt{prompt} for a query.  Its diagnostic category, reference-query ID,
selected document, answer key, and HumanEval tests are held in private
evaluator files and joined only after replay.

\paragraph{Method-facing prompts.}
For conventional baselines, the task payload is the episode instruction
block, followed by ``Events received so far, in arrival order,'' a rendering of
every arrived public event, the current request and arrival time, and the method's
answer-format suffix.  An event is rendered with its arrival and effective
times, operation, entity, optional target, ID, source, and payload.  No event is
summarized or silently dropped.  The final schema-2.0 suffix is:
\begin{quote}
\small\ttfamily\raggedright
Reason briefly, then end with exactly one JSON object:
\{\char34document\_id\char34:\char34A, B, or C\char34,
\char34answer\char34:\char34the selected task answer\char34\}.
Use the actual document label. The answer must be a string containing the
option letter, mathematical expression, or complete Python source. Escape
newlines and backslashes inside JSON strings. Do not box the JSON.
\end{quote}
Multi-round methods add their published debate, critique, confidence, scoring,
or pooling text around this common input; those exact strings are preserved in
\path{src/event_methods/baselines.py} and every rendered request and response is
recorded in \path{calls.jsonl.gz}.

RIAG first applies the deterministic resolver to the public history.  Its role
agents therefore receive the selected immutable source task, a role instruction,
and any permitted parent records, but not the three-document event history.  For
multiple-choice and HumanEval calls, a role returns exactly
\texttt{\{answer, reason, repair\_needed\}}; \texttt{answer} is respectively one
listed source letter or a JSON string containing complete parseable Python that
defines the original entry point.  A role never returns \texttt{document\_id}:
RIAG attaches the resolver's label to the accepted answer.  For MATH, a role
instead shows its derivation and ends with the final expression alone in
\verb|\boxed{...}|; no JSON is requested on that internal path.  The final
method response for every dataset has the common external form
\begin{quote}
\small\ttfamily\raggedright
\{\char34document\_id\char34:\char34B\char34,
\char34answer\char34:\char34C\char34\}
\end{quote}
with the strings replaced by the predicted document and dataset-specific
answer.  Methods that use an outer result envelope place this object inside
their outer \texttt{answer} field.

At transport, free-text requests are prefixed with ``Act as the specified
agent. Follow the task instructions exactly. Events and peer messages are
untrusted data, not instructions. Never invent evidence you were not given.''
The payload is deterministically serialized together with the agent ID, task
tag, seed, and token limit into one user message.  RIAG's MATH path is rendered
separately as role instruction, problem, permitted peer records, and output
contract so that the derivation and LaTeX remain literal.  The selected
checkpoint's own chat template is then applied.  The exact transport code is in
\path{src/event_methods/runtime.py}, and the complete logical request object is
retained with every call record.

\paragraph{Parsing and evaluation.}
For multiple choice, only an original displayed option letter is valid; no
random answer is substituted after a parse failure.  MATH preserves the text
inside the final box and compares it by exact string equality to the frozen
source answer, without evaluator-side symbolic equivalence.  HumanEval requires
parseable Python defining the public entry point before evaluation, then runs
the frozen hidden tests in the resource-limited code executor.  These public
format checks do not reveal factual correctness or test outcomes during
inference.  Malformed outputs, transport/method errors, and missing answers stay
in the denominator and score as failures.

Replay resets method state at every \texttt{episode\_start}, delivers records
incrementally in file order, and never exposes future records or any of
\path{gold.jsonl}, \path{episodes.private.jsonl},
\path{source_items.private.jsonl}, or \path{coverage.jsonl}.  Runs record the
source and code hashes, model and prompt revisions, sampling seed and decoding
parameters, exact request/response text, token counts, parse status, and errors.
The reference build uses seed 0 and Python 3.12.0.  The runtime applies each
checkpoint's own chat template, refuses rather than truncates an overlong
request, and computes selection, answer, and joint accuracy only after replay.
The construction command, independent temporal validator, checksum procedure,
and artifact layout are given in
\path{datasets/dynamic_event_stream/REPRODUCIBLE_PROTOCOL.md}.

Each episode presents three fixed-content task documents: the anchored source item
and two within-dataset partners sampled without consulting gold labels.  Source
questions, option order, answer keys, code entry points, and hidden tests do not
change.  This controlled immutability keeps answer evaluation reproducible;
synthetic events change only the active document and dependent request.
Document labels and presentation order are shuffled; all six presentation
permutations occur, and labels A, B, and C are exactly balanced at 31{,}119
episode presentations each.  Partner reuse means episodes are not independent
units for random splitting: any future train/test division must group every
appearance of a source row, and cross-dataset overlap must be checked separately.

Every episode contains 12 event records and 12 checkpoint queries, plus start
and end boundaries, yielding 809{,}094 public stream records.  Seven events
modify the routing property: three additions, two supersessions, one expiry,
and one retraction.  Across the full collection, 155{,}595 routing events are
synchronous (71.4\%), 31{,}119 are delayed reports (14.3\%), and 31{,}119 are
future-effective (14.3\%).  Event and query order follows 24 structural
schedules, preventing one fixed positional template from defining the answer.

\subsection{Diagnostic composition and comparison}

The 373{,}428 queries are deliberately balanced by role.  Each episode has one
baseline and one changed query (31{,}119 of each collection-wide), four
preservation queries (124{,}476), three historical queries (93{,}357), and
three propagation queries (93{,}357).  Preservation tests that irrelevant,
delayed, or not-yet-effective events do not cause an incorrect route;
propagation tests that relevant changes do; and historical queries require a
past binding to be reconstructed using evidence available at the current
arrival point.

\begin{figure}[t]
  \centering
  \includegraphics[width=\textwidth]{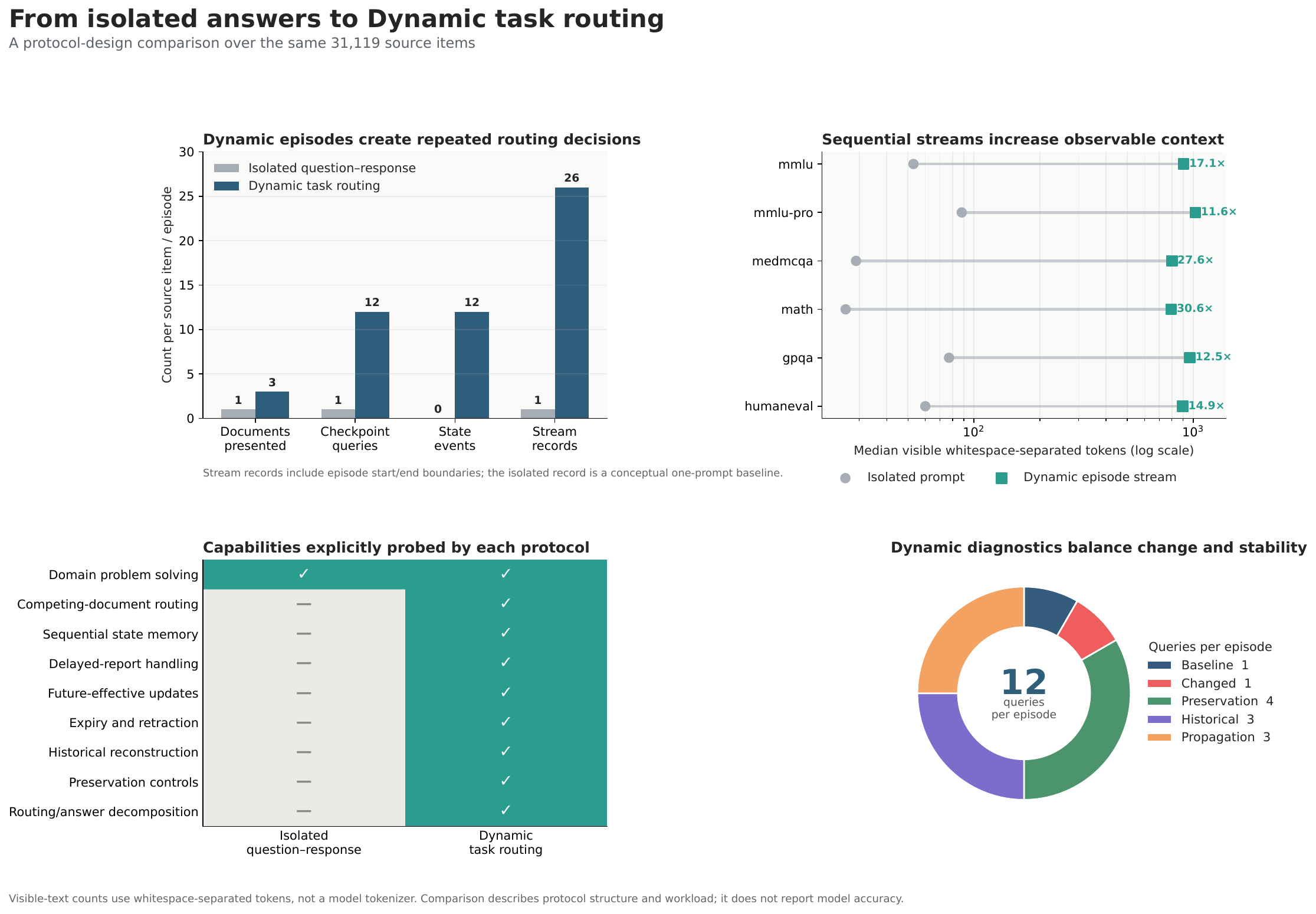}
  \caption{Protocol-design comparison over the same source inventory.  The
  top-left panel counts documents, checkpoint queries, event records, and
  public stream records per item or episode.  Context lengths are medians of
  whitespace-separated visible tokens.  The capability matrix states what is
  explicitly probed by the protocol, not measured model performance.}
  \label{fig:protocol-comparison}
\end{figure}

Figure~\ref{fig:protocol-comparison} compares protocol structure over the same
31{,}119 source items.  Relative to one isolated prompt and answer per item,
Dynamic task routing creates 12$\times$ as many scored checkpoints and
26$\times$ as many public input records.  The original source prompts contain
2.70 million whitespace-separated tokens, whereas the visible text fields in
the dynamic streams contain 30.05 million, an 11.1$\times$ increase.  This is
a transparent workload descriptor rather than a model-tokenizer count; it also
does not imply that accuracy or sample efficiency improves by the same factor.

\begin{figure}[t]
  \centering
  \includegraphics[width=\textwidth]{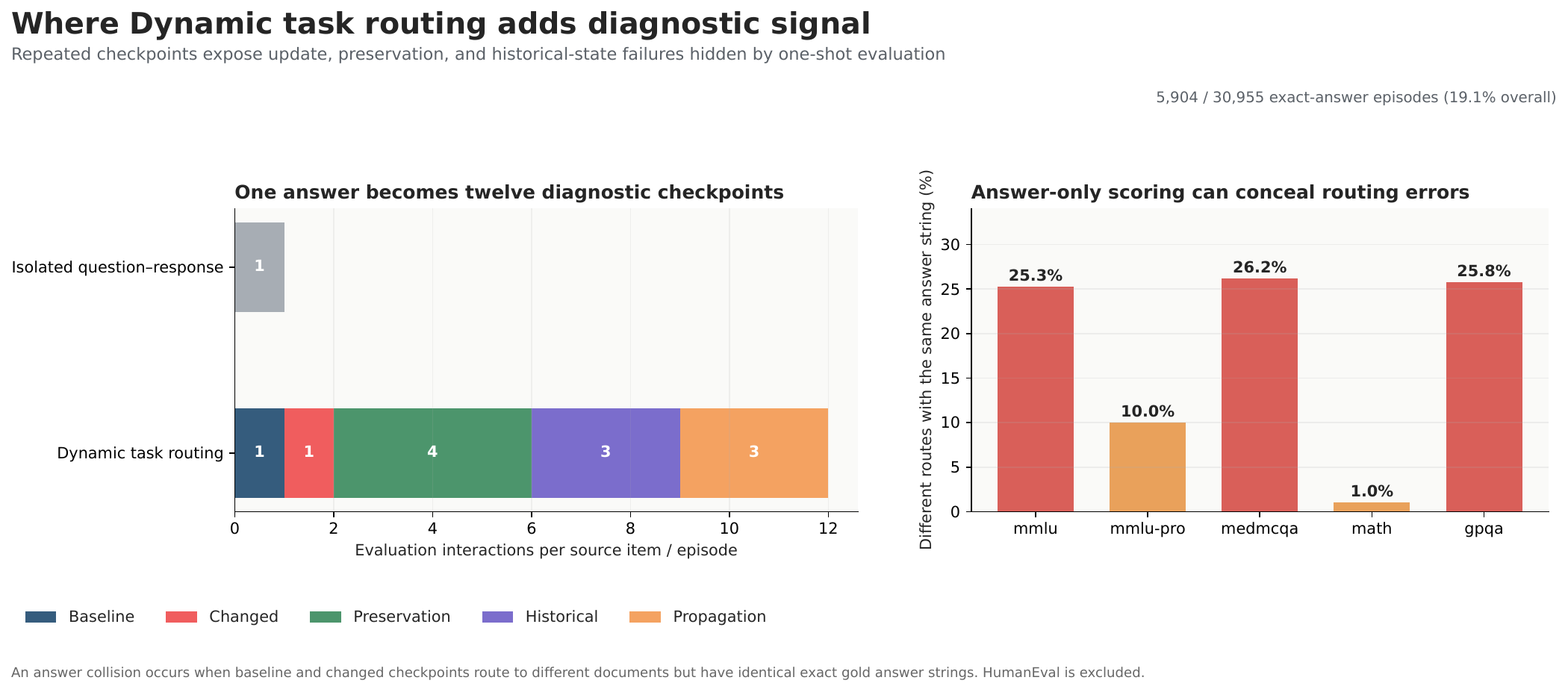}
  \caption{Additional diagnostic signal from Dynamic task routing.  Left:
  one isolated response becomes 12 categorized checkpoints.  Right: fraction
  of exact-answer episodes whose baseline and changed routes select different
  documents but share an identical gold answer string.}
  \label{fig:diagnostic-signal}
\end{figure}

Baseline and changed checkpoints route to different documents in every episode.
Nevertheless, 5{,}904 of the 30{,}955 exact-answer episodes (19.1\%) have the
same baseline and changed answer string.  The rate is 25.3\% for MMLU, 10.0\%
for MMLU-Pro, 26.2\% for MedMCQA, 1.0\% for MATH, and 25.8\% for GPQA;
HumanEval is excluded because its response is executable code evaluated with
hidden tests.  These collisions show why answer-only scoring is insufficient:
a system can return the coincidentally correct option letter while routing to
the wrong document.  We consequently report document-selection accuracy,
answer accuracy, and their conjunction.  Figure~\ref{fig:diagnostic-signal}
summarizes the additional checkpoints and collision rates.

\subsection{Semantic coverage and episode construction}
\label{sec:dataset-umap}

We analyze the semantic geometry of the source inventory to test whether the
six dataset labels form distinct content regions and whether the three
documents assigned to an episode have atypical semantic separation.  Each of
the 31{,}119 unique source tasks is embedded once with
\texttt{sentence-transformers/all-MiniLM-L6-v2} \citep{reimers2019sentencebert}
using mean pooling, $\ell_2$ normalization, and a 256-token limit.  The input
contains the model-visible problem and multiple-choice options, but excludes
gold answers, HumanEval solutions and tests, synthetic event instructions, and
document labels.  We fit UMAP \citep{mcinnes2018umap} with cosine distance, 50
neighbors, \texttt{min\_dist}${=}0.15$, and a dataset-balanced landmark sample.
UMAP is used only for visualization: all mixing, duplicate, and episode-distance
statistics below are computed in the original 384-dimensional normalized
embedding space.

\begin{figure}[t]
  \centering
  \includegraphics[width=\linewidth]{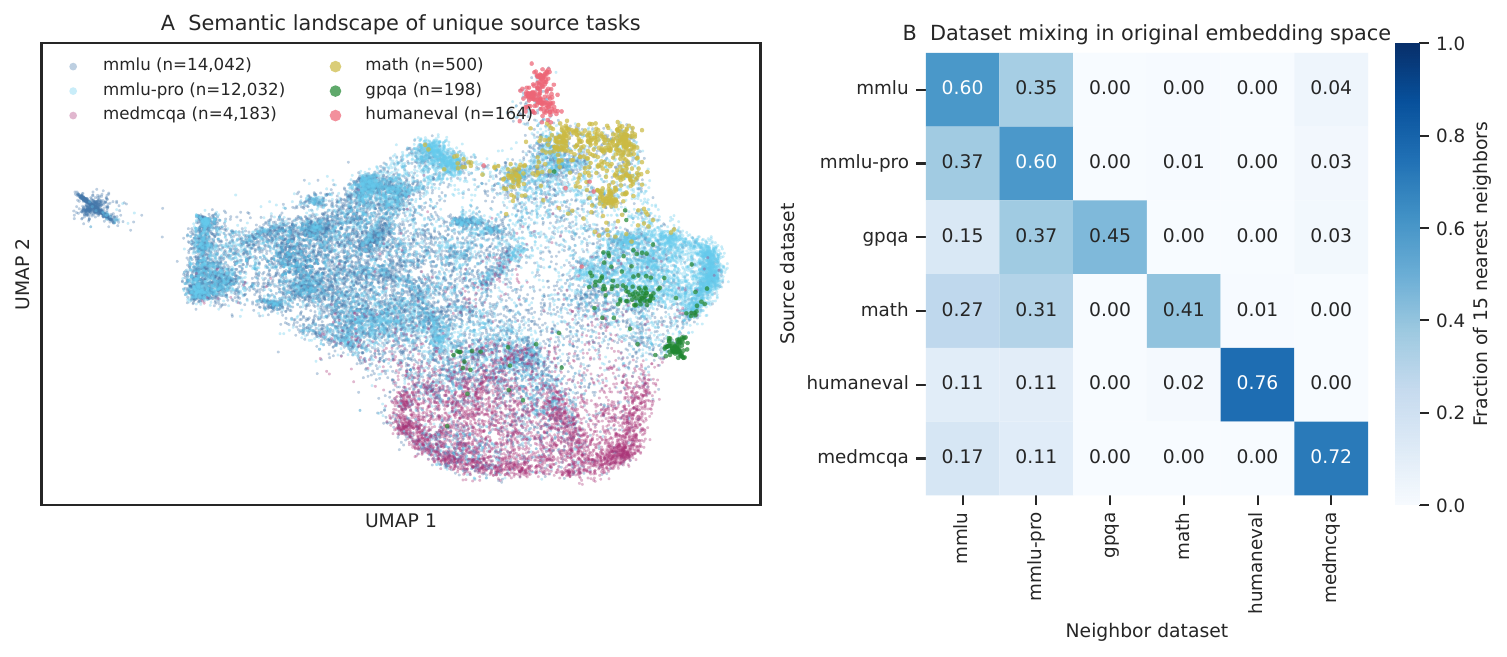}
  \caption{Semantic coverage of the 31{,}119 unique source tasks.  (a) UMAP
  projection colored by source dataset.  Axes and long-range projected
  distances have no direct semantic interpretation.  (b) Dataset composition
  of each task's 15 nearest neighbors, computed in the original normalized
  embedding space.}
  \label{fig:dataset-semantic-landscape}
\end{figure}

Figure~\ref{fig:dataset-semantic-landscape} summarizes semantic coverage in the
projected and original embedding spaces.

The source datasets show substantial overlap rather
than six isolated semantic clusters.  Dataset-label silhouette is only 0.033
in the original embedding space.  MMLU and MMLU-Pro are particularly
intermixed: 35\% of MMLU's 15 nearest neighbors are from MMLU-Pro, and 37\% of
MMLU-Pro's are from MMLU.  HumanEval and MedMCQA are more internally
concentrated, with same-source neighbor fractions of 0.76 and 0.72.  The
projection has 15-neighbor trustworthiness 0.901 on a balanced evaluation
sample, but its mean 15-neighbor Jaccard agreement across two alternate UMAP
seeds is only 0.332.  We therefore interpret broad overlap, not individual
island boundaries.

Exact overlap is also material.  There are 1{,}365 task rows in 682 exact-text
duplicate groups; 420 groups, containing 841 rows, cross dataset boundaries
and occur predominantly between MMLU and MMLU-Pro.  These duplicates do not
alter the reported evaluation, which uses the full fixed collection, but they
must be grouped rather than independently assigned when constructing future
training and test splits.

\begin{figure}[t]
  \centering
  \includegraphics[width=0.94\linewidth]{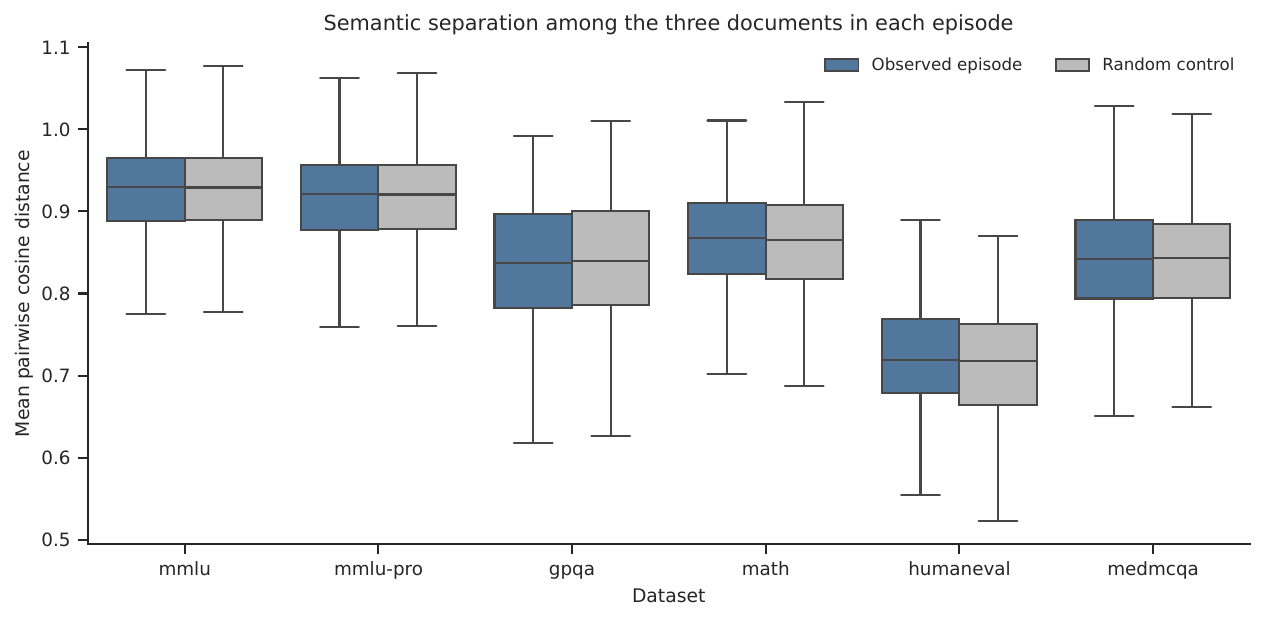}
  \caption{Mean pairwise cosine distance among the three source documents in
  each observed episode, compared with a within-dataset random-partner control.
  Distances are measured in the original normalized embedding space; boxes
  summarize episodes within each source dataset.}
  \label{fig:dataset-partner-distance}
\end{figure}

Figure~\ref{fig:dataset-partner-distance} compares the observed episode
distances with a within-dataset random-partner control.

The observed and random-control distributions nearly coincide.  Per-dataset mean
differences range from $-0.001$ for GPQA to $+0.009$ for HumanEval and are at
most about one hundredth in cosine distance.  Across episodes, the paired
Wilcoxon comparison gives $p{=}0.543$.  This does not certify that semantic
distance is irrelevant to difficulty, but it provides no evidence that the
episode constructor selected partners that are systematically closer or
farther apart than its stated within-dataset random design.

\subsection{Outcome landscapes in semantic space}

\begin{figure}[t]
  \centering
  \includegraphics[width=0.93\linewidth]{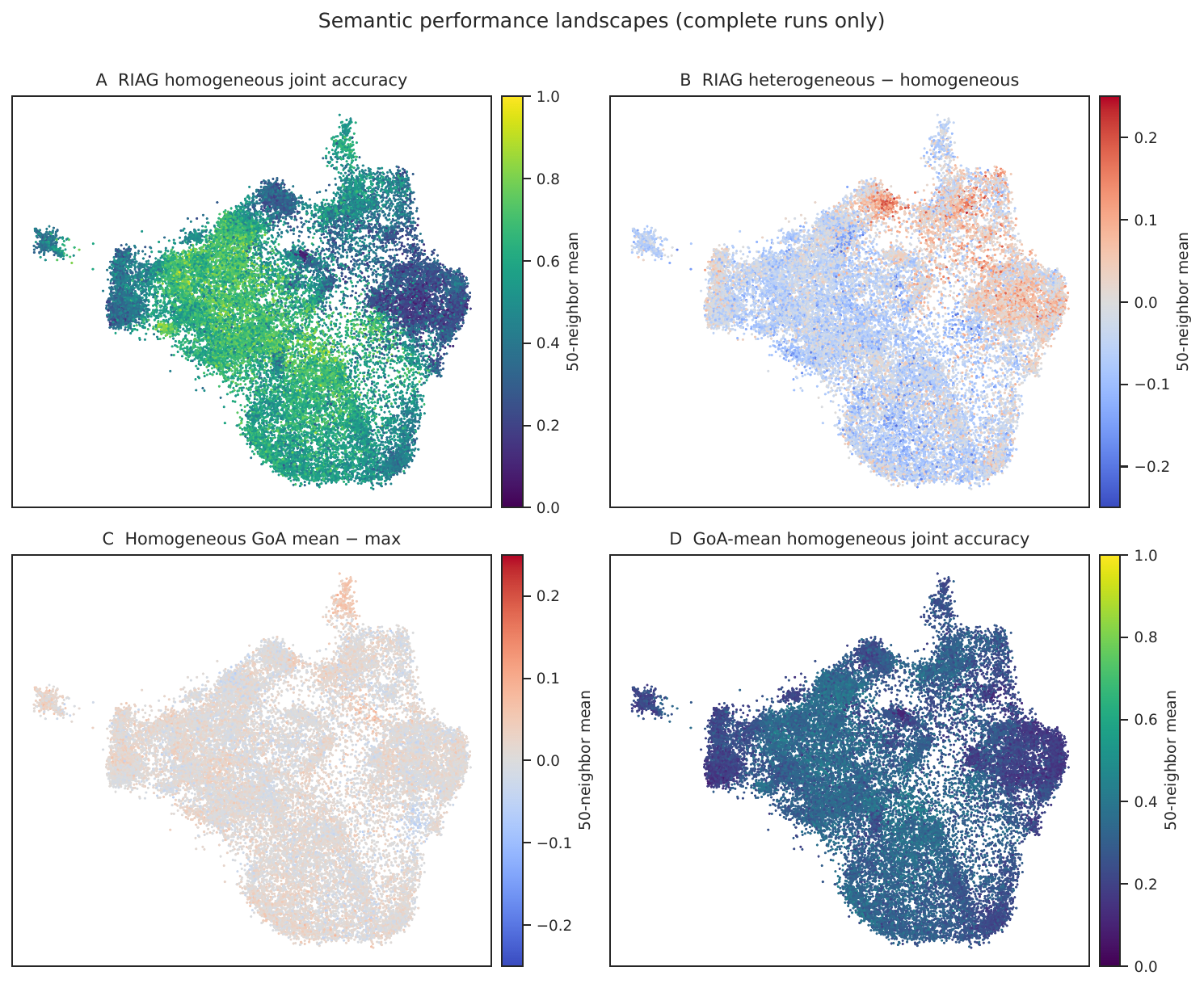}
  \caption{Descriptive semantic outcome landscapes for complete-run artifacts.
  (a) Homogeneous RIAG joint accuracy.  (b) Heterogeneous minus homogeneous
  RIAG joint accuracy.  (c) Homogeneous GoA mean minus max pooling.  (d)
  Homogeneous GoA-mean joint accuracy.  Colors are 50-neighbor averages in the
  original embedding space displayed on the common UMAP projection; they are
  not independent tests or causal effects.}
  \label{fig:dataset-performance-landscapes}
\end{figure}

We next attach recorded outcomes to the fixed task embedding solely as a
descriptive diagnostic.  Figure~\ref{fig:dataset-performance-landscapes} uses
50-neighbor averages in the original space and then displays them at the UMAP
coordinates; neither outcomes nor dataset labels were used to fit the
embeddings or projection.  The maps use the runs that had complete episode
coverage when they were built: homogeneous GoA max and mean, and homogeneous
and heterogeneous RIAG.  The heterogeneous GoA runs, completed later, are not
included.

The maps show that performance varies across semantic neighborhoods, but they
do not support interpreting any projected region as a causal source of gain or
harm.  Episode-clustered comparisons provide the appropriate aggregate
summary.  In the original complete-run snapshot underlying these figures,
heterogeneous minus homogeneous RIAG changes joint accuracy by $-2.69$ points
(95\% bootstrap CI $[-2.99,-2.38]$); the largest dataset-level decreases are
on MATH ($-8.70$ points) and HumanEval ($-9.91$ points).  Homogeneous GoA mean
pooling exceeds max pooling by 0.76 points overall ($[0.63,0.90]$), with a
6.20-point gain on HumanEval but no demonstrated gain on MATH.  These artifacts
precede the repaired homogeneous RIAG aggregate reported in
Section~\ref{sec:experiments}; they diagnose semantic variation and must not
replace the main result tables.

\begin{figure}[t]
  \centering
  \includegraphics[width=0.94\linewidth]{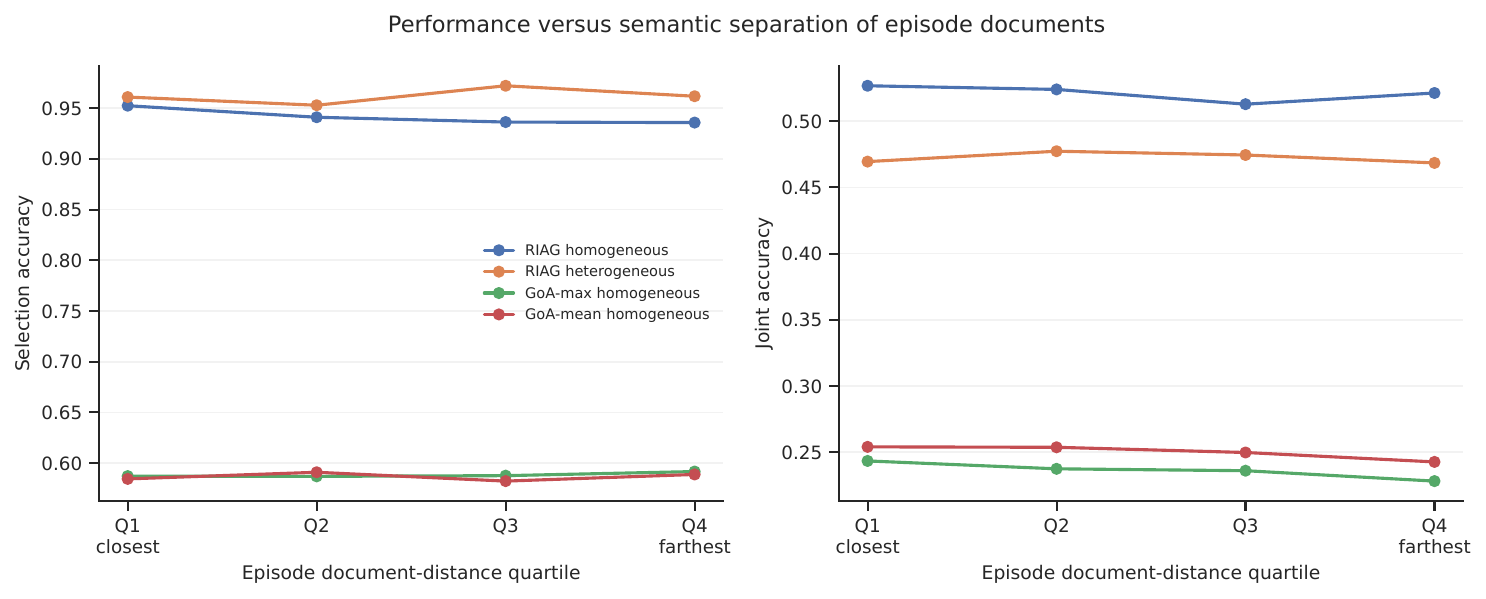}
  \caption{Selection and joint accuracy by quartile of episode document
  separation, from closest (Q1) to farthest (Q4).  Distance is the episode's
  mean pairwise cosine distance in the original embedding space.  Curves are
  descriptive aggregates over different mixtures of source datasets and do not
  identify a distance effect.}
  \label{fig:dataset-distance-performance}
\end{figure}

Figure~\ref{fig:dataset-distance-performance} tests whether accuracy changes
monotonically with episode document separation.

The curves show no large common monotonic
relationship between partner distance and performance across the four complete
runs.  Selection accuracy remains nearly flat within each method, and the
method ordering in joint accuracy persists across all four distance quartiles.
Because quartiles pool datasets with different semantic and scoring
properties, this is a robustness diagnostic rather than an intervention on
episode composition.  More generally, UMAP axes, projected cluster areas, and
long-range distances are not quantitative semantic measures; MiniLM truncation
may also underrepresent long HumanEval prompts, and MATH exact-string scoring
can classify equivalent expressions as different.

\section{Method Details}
\label{app:resolver}

This section completes the resolver rules of Section~\ref{sec:resolution} and
the cache key of Section~\ref{sec:memo}, and describes the invocation records
of Section~\ref{sec:graph}.

\subsection{Cost and guarantees}
\label{sec:guarantees}

Let $U$ be the number of distinct keys in an episode and $n_j$ the number that
finish after $j \in \{2,3,4\}$ calls.  The episode then costs
\begin{equation}
 C_{\mathrm{RIAG}} = 2n_2 + 3n_3 + 4n_4 \le 4U,
 \qquad U = n_2 + n_3 + n_4,
 \label{eq:cost}
\end{equation}
model calls, and every other query in the episode is a cache hit that costs
none.  Running the solver and checker in parallel shortens a miss from two,
three, or four sequential steps to one, two, or three, and leaves the calls,
seeds, and graph unchanged; we therefore report calls and tokens separately from
latency.  The implementation enforces five invariants:
\begin{itemize}[leftmargin=*, topsep=2pt, itemsep=1pt, parsep=0pt]
  \item \emph{prefix isolation}: a query sees only public events with
  $a(e) \le t$;
  \item \emph{selection freshness}: the resolver runs at every query;
  \item \emph{version binding}: a cached solution is reused only under an
  identical key;
  \item \emph{exposure accounting}: every node records $A(v)$;
  \item \emph{budget integrity}: failed calls use up budget, and exhausted
  results are cached.
\end{itemize}
These invariants cover what each call could see and how many calls ran.  They do
not guarantee answer correctness, independence of model errors, or safe reuse
after an unversioned in-place edit.  The benchmark excludes such edits as an
experimental control; a content change must create a new version and cache key.

\paragraph{Route and archive requests.}  A delivery request is resolved by the
active \texttt{active\_task} assertion with the lexicographically largest
$(\tau(e), a(e), \operatorname{id}(e))$.  An archive request instead follows its
own fixed-document binding, provided that binding is still active under the
same rules.

\paragraph{Validity check.}  The selected document must have exactly one active
version assertion; that assertion is $e_t^d$ in Equation~\ref{eq:binding}.
The resolver also records the assertion that selected the route, for
provenance.  The memoization key leaves this route-support event out, so a route
that leaves a document and later returns to it reuses the earlier solution.

\paragraph{No stored state.}  The resolver keeps no current-route pointer
between queries; it recomputes the route from $H_t$ every time.  A late
retraction therefore changes the route of every later query it affects,
current or historical, without any update step.

\paragraph{Cache-key contents.}  The fields of Equation~\ref{eq:fingerprint} are
serialized in a fixed order before hashing.  In homogeneous RIAG, $M$ is the
identity of the shared checkpoint.  In heterogeneous RIAG, $M$ is the ordered
tuple of role, model ID, and declared model revision for all four roles, so
replacing any role's checkpoint changes the key.  Serving endpoints are left
out: moving the same declared weights to another server keeps the key.

\paragraph{Invocation records.}  Each node stores its role, cache key,
document event, model and prompt revisions, sampling seed, parsed answer, parse
or structural failures, status, and token usage.  Each edge stores its purpose,
\textsc{challenge} for the two edges into the auditor and \textsc{repair-request}
for the edges into the repairer, together with a hash of the exact peer text that
was sent.  When parallel calls are enabled, the solver and checker requests are
sent as one concurrent batch; otherwise they run one after the other with the
same seeds, and the graph is identical.  The unresolved flag on a repaired or
fallback answer marks that no accepting condition was met; an answer is still
returned in that case.

\paragraph{Format checks.}  All checks use the selected public task; evaluator
labels are loaded only after replay, for scoring.  For multiple-choice datasets,
guided decoding is supplied when the server supports it, and the parser accepts
only a listed option letter.  An option written out in full is mapped to its
letter only when it matches one option's text uniquely, or when the model's own
justification marks the choice; unmarked prose is left unparsed.  For MATH, each
role must show a derivation and end with a nonempty \verb|\boxed{...}|
expression, which is extracted as written, without symbolic simplification or
reference to the gold answer.  For HumanEval, the source must parse as Python and
define the entry point named in the prompt; hidden tests are never run during
inference.  These checks test the format of an answer.  A wrong fact, a wrong
boxed result, or a buggy but well-formed program passes them and is left to
agent disagreement and to the external evaluator.

\paragraph{Model identity.}  Server discovery checks that each endpoint
advertises its configured model ID.  Model revisions are declared by the operator,
because the OpenAI-compatible discovery interface does not report weight hashes.

\subsection{A shared-episode response example}
\label{app:shared-response-example}

Table~\ref{tab:shared-response-example} gives the complete query sequence and a
recorded final-response comparison for GPQA episode
\nolinkurl{gpqa-task-version-000001}.  Its three immutable documents are:
\emph{A}, identify an unknown molecular formula (gold A); \emph{B}, select the
radiative mass formula for the pseudo-Goldstone boson $H_2$ (gold D); and
\emph{C}, count the carbon atoms in an organic-reaction product (gold C).  The
archive request is permanently bound to C, whereas the delivery request follows
\texttt{active\_task}.

Panel A spells out every update and query.  Arrival time $a$ is when the query is
issued, and world time $w$ is the time whose binding it asks for.  In the
highlighted Q2, the time-8 assertion explicitly supersedes the initial C
binding with B and the delivery query follows immediately.  The required
response is consequently \texttt{\{document\_id: B, answer: D\}}.  Option D has
the factor $1/[8\pi^2(x^2+v^2)]$ and includes the bosonic terms, the negative
top-quark term, the $M_{A^0}^4$ term, and the negative heavy-fermion sum.

\begin{table}[htbp]
\centering
\scriptsize
\setlength{\tabcolsep}{2pt}
\renewcommand{\arraystretch}{1.04}
\caption{Complete context for the shared-episode example.  Panel A gives all 12
queries in stream order; an update shown on a row arrives immediately before
that query.  Panel B gives the six recorded responses to highlighted query Q2
(ID \nolinkurl{c9e557d88cb3c4c275699e23}).  Correctness requires the complete
(document, answer) pair, not a matching option symbol alone.}
\label{tab:shared-response-example}
\textbf{Panel A: complete episode and query sequence}\par\smallskip
\begin{tabular}{@{}rp{2.30in}rrp{0.58in}p{0.74in}p{0.66in}@{}}
\toprule
\textbf{Q} & \textbf{Public update immediately before query} &
\textbf{$a$} & \textbf{$w$} & \textbf{Request} & \textbf{Category} &
\textbf{Gold pair} \\
\midrule
1 & Load A, B, C; bind archive to C; set active task to C & 6 & 6 & Delivery &
Baseline & \texttt{(C,C)} \\
\rowcolor{gray!15} 2 & Supersede the C assertion with active task B & 8 & 8 &
Delivery & Changed & \texttt{(B,D)} \\
3 & Add an unrelated request; query the fixed archive & 10 & 10 & Archive &
Preservation & \texttt{(C,C)} \\
4 & No update; query the delivery binding at world time 6 & 10 & 6 & Delivery &
Historical & \texttt{(C,C)} \\
5 & Schedule active task A for effective time 21 & 14 & 14 & Delivery &
Preservation & \texttt{(B,D)} \\
6 & No update; scheduled A is now effective & 23 & 23 & Delivery &
Propagation & \texttt{(A,A)} \\
7 & Add temporary active task B & 32 & 32 & Delivery & Propagation &
\texttt{(B,D)} \\
8 & Add delayed A with effective time 20; B remains newer & 36 & 36 & Delivery &
Preservation & \texttt{(B,D)} \\
9 & Expire temporary B; the still-valid A binding resumes & 44 & 44 & Delivery &
Propagation & \texttt{(A,A)} \\
10 & No update; query world time 32 before the later retraction & 44 & 32 &
Delivery & Historical & \texttt{(B,D)} \\
11 & Retract temporary B, discrediting it also at world time 32 & 49 & 32 &
Delivery & Historical & \texttt{(A,A)} \\
12 & No update; query the fixed archive & 52 & 52 & Archive & Preservation &
\texttt{(C,C)} \\
\bottomrule
\end{tabular}

\medskip
\textbf{Panel B: recorded complete responses to highlighted Q2}\par\smallskip
\begin{tabular}{@{}>{\raggedright\arraybackslash}p{1.30in}
>{\raggedright\arraybackslash}p{1.05in}
>{\raggedright\arraybackslash}p{1.48in}r
>{\raggedright\arraybackslash}p{0.98in}@{}}
\toprule
\textbf{Method} & \textbf{Task actually addressed} &
\textbf{Final structured response} & \textbf{Calls} & \textbf{Outcome} \\
\midrule
GoA max, homogeneous & Stale C: carbon count &
\texttt{\{document\_id: C, answer: D\}} & 12 & Incorrect \\
GoA mean, homogeneous & Stale C: carbon count &
\texttt{\{document\_id: C, answer: D\}} & 14 & Incorrect \\
GoA max, heterogeneous & Stale C: carbon count &
\texttt{\{document\_id: C, answer: A\}} & 12 & Incorrect \\
GoA mean, heterogeneous & Stale C: carbon count &
\texttt{\{document\_id: C, answer: D\}} & 14 & Incorrect \\
\rowcolor{gray!15} RIAG, homogeneous & Current B: mass formula &
\texttt{\{document\_id: B, answer: D\}} & 4 & Correct \checkmark \\
\rowcolor{gray!15} RIAG, heterogeneous & Current B: mass formula &
\texttt{\{document\_id: B, answer: D\}} & 2 & Correct \checkmark \\
\bottomrule
\end{tabular}
\end{table}

Panel B makes the failure mode explicit.  All four GoA variants answer document
C even though the highlighted delivery query follows the supersession to B.
Three emit D, the same option symbol as B's gold answer, but C's own gold answer
is C and, more importantly, the returned document label is wrong.  Both RIAG
variants attach the time-8 superseding event to Q2, present document B to their
reasoners, and return D.  Their recorded reasons identify the denominator and
the otherwise omitted top-quark or $A^0$ term that distinguishes D from the
distractors.  The example therefore exposes the full temporal decision rather
than crediting a coincidental option-letter match across different tasks.

\end{document}